\documentclass[journal]{IEEEtran}

\usepackage{graphicx}
\usepackage{amssymb,amsmath,epsfig}
\usepackage{algorithm}
\usepackage{algorithmic}
\usepackage{booktabs}
\usepackage{mathrsfs}
\usepackage{color}
\usepackage{textcomp}
\usepackage{bbding}
\usepackage{pifont}
\usepackage{wasysym}
\usepackage{amssymb}
\usepackage{subcaption}
\usepackage{cite}
\usepackage{amsmath,amssymb,amsfonts}
\usepackage{xcolor}
\usepackage{multirow}
\usepackage{epsfig}
\usepackage[hyphens]{url}
\usepackage[colorlinks,
            linkcolor=red,
            anchorcolor=blue,
            citecolor=green
            ]{hyperref}
\begin{document}
%
\title{CTCNet: A CNN-Transformer Cooperation Network \\ for Face Image Super-Resolution }
%
%
%
\author{Guangwei Gao,~\IEEEmembership{Senior Member,~IEEE,}
        Zixiang Xu,
        Juncheng Li,
        Jian Yang,~\IEEEmembership{Member,~IEEE,}
        \\ Tieyong Zeng,~\IEEEmembership{Member,~IEEE,}
        and Guo-Jun~Qi,~\IEEEmembership{Fellow,~IEEE}
\thanks{This work was supported in part by the National Natural Science Foundation of China under Grants 61972212 and 61833011, the Six Talent Peaks Project in Jiangsu Province under Grant RJFW-011, and the Open Fund Project of Provincial Key Laboratory for Computer Information Processing Technology (Soochow University) under Grant KJS2274.~\textit{(Guangwei Gao and Zixiang Xu contributed equally to this work.) (Corresponding author: Juncheng Li)}}
\thanks{Guangwei Gao and Zixiang Xu are with the Institute of Advanced Technology, Nanjing University of Posts and Telecommunications, Nanjing, China, and also with the Provincial Key Laboratory for Computer Information Processing Technology, Soochow University, Suzhou, China (e-mail: csggao@gmail.com, wszixiangxu@gmail.com).}
\thanks{Juncheng Li is with the School of Communication \& Information Engineering, Shanghai University, Shanghai, China, also with Jiangsu Key Laboratory of Image and Video Understanding for Social Safety, Nanjing University of Science and Technology, Nanjing, China (e-mail: cvjunchengli@gmail.com).}
\thanks{Jian Yang is with the School of Computer Science and Technology, Nanjing University of Science and Technology, Nanjing, China (e-mail: csjyang@njust.edu.cn).}
\thanks{Tieyong Zeng is with the Center for Mathematical Artificial Intelligence, Department of Mathematics, The Chinese University of Hong Kong, Hong Kong, China (e-mail: zeng@math.cuhk.edu.hk).}
\thanks{Guo-Jun Qi is with Research Center for Industries of the Future and School of Engineering at Westlake University, and OPPO Research Seattle, USA (e-mail: guojunq@gmail.com).}
}


\markboth{IEEE Transactions on Image Processing}%
{Shell \MakeLowercase{\textit{et al.}}: Bare Demo of IEEEtran.cls for IEEE Journals}

%

\maketitle

\begin{abstract}
Recently, deep convolution neural networks (CNNs) steered face super-resolution methods have achieved great progress in restoring degraded facial details by joint training with facial priors. However, these methods have some obvious limitations. On the one hand, multi-task joint learning requires additional marking on the dataset, and the introduced prior network will significantly increase the computational cost of the model. On the other hand, the limited receptive field of CNN will reduce the fidelity and naturalness of the reconstructed facial images, resulting in suboptimal reconstructed images. In this work, we propose an efficient CNN-Transformer Cooperation Network (CTCNet) for face super-resolution tasks, which uses the multi-scale connected encoder-decoder architecture as the backbone. Specifically, we first devise a novel Local-Global Feature Cooperation Module (LGCM), which is composed of a Facial Structure Attention Unit (FSAU) and a Transformer block, to promote the consistency of local facial detail and global facial structure restoration simultaneously. Then, we design an efficient Feature Refinement Module (FRM) to enhance the encoded features. Finally, to further improve the restoration of fine facial details, we present a Multi-scale Feature Fusion Unit (MFFU) to adaptively fuse the features from different stages in the encoder procedure. Extensive evaluations on various datasets have assessed that the proposed CTCNet can outperform other state-of-the-art methods significantly. Source code will be available at \textit{\url{https://github.com/IVIPLab/CTCNet}}.
\end{abstract}


%
\IEEEpeerreviewmaketitle

\section{Introduction}
\label{sec1}

\IEEEPARstart{F}{ace} super-resolution (FSR), a.k.a. face hallucination, refers to a technology for obtaining high-resolution (HR) face images from input low-resolution (LR) face images. In practical application scenarios, due to the inherent differences in the hardware configuration, placement position, and shooting angle of the image capture device, the quality of the face images obtained by shooting is inevitably poor. Lower-quality images seriously affect downstream tasks such as face analysis and face recognition. Unlike general image super-resolution, the core goal of FSR is to reconstruct as much as possible the facial structure information (i.e., shapes of face components and face outline) that is missing in the degraded observation. Although these structures only occupy a small part of the face, they are the key to distinguishing different faces. Compared with other areas in a face image, the facial feature and contours of a person are usually more difficult to restore since they often span a large area and require more global information.

Most of the previous FSR algorithms~\cite{ma2020deep,hu2020face,cai2019fcsr} mainly adopted the strategy of successive multi-task training. These methods used facial landmark heatmaps or parsing maps to participate in the formal training to constrain the performance of the FSR reconstruction network. However, they also need extra labeled data to achieve the goal. Besides, in the previous FSR methods~\cite{chen2018fsrnet,xin2020facial}, the encoding and decoding parts are connected in series. This kind of connection cannot fully utilize the low-level features, and the low-level features also cannot thoroughly guide the learning of the high-level features, resulting in unsatisfied performance in the FSR task. 
In addition, many FSR networks~\cite{zhang2018super,kim2019progressive,chen2020learning,yang2021gan,wang2022restoreformer} have been built using Convolution Neural Networks (CNNs) due to the powerful local modeling capabilities of CNN to predict fine-grained facial details. However, the human face usually has a fixed geometric feature structure~\cite{chen2020robust,gao2021constructing,zhang2022pro}. Therefore, if only focusing on extracting the local information while ignoring the relationship between them (global information), it will inevitably affect the restoration of the global facial structure, leading to blurry effects in the generated faces.

As we all know, local methods (such as CNN-based methods) mainly focus on the local facial details, while global methods (such as Transformer-based methods) usually capture the global facial structures. How to collaboratively make full use of the local and global features, and how to efficiently aggregate the multi-scale abundant features is important. To achieve this, in this work, we propose an efficient CNN-Transformer Cooperation Network (CTCNet) for FSR. Like most previous FSR models, our CTCNet also uses an encoder-decoder structure. Specifically, in the encoder and decoder branches, the specially designed Local-Global Feature Cooperation Module (LGCM) is used for feature extraction. LGCM comprises a Facial Structure Attention Unit (FSAU) and a Transformer block. Among them, FSAU is specially designed to extract key face components information, and Transformer blocks are introduced to explore long-distance visual relation modeling. The combination of FASU and Transformer block can simultaneously capture local facial texture details and global facial structures. Meanwhile, instead of using successive connections, we design a Multi-scale Feature Fusion Unit (MFFU) to fuse the features from different stages of the network flexibly. In addition, we use the Feature Refinement Modules (FRMs) between the encoder and decoder branches to further enhance the extracted features, thus improving the performance of CTCNet. In summary, the main contributions of this work are as follows

\begin{figure*}[t]
	\centerline{\includegraphics[width=18cm]{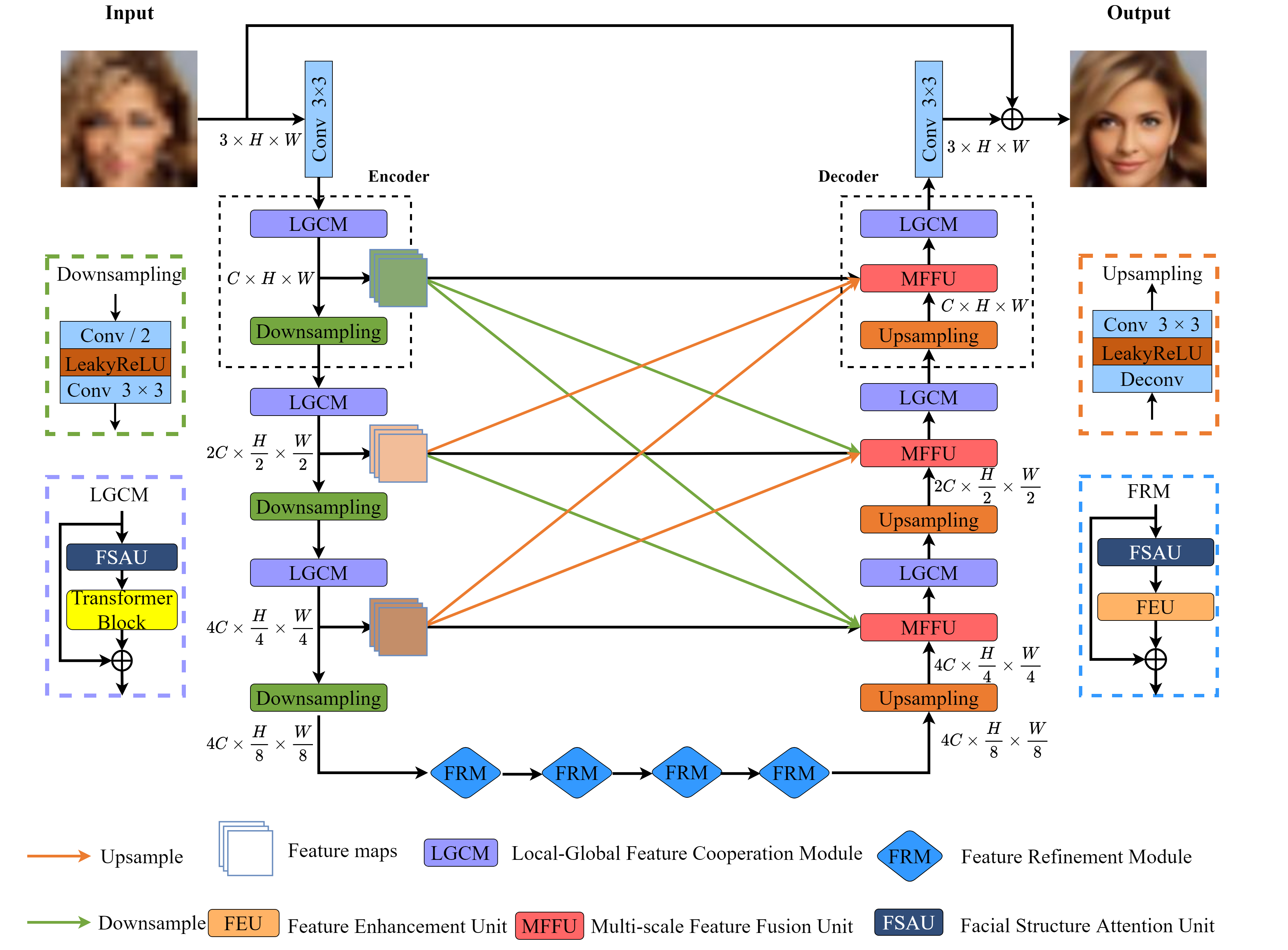}}
	\caption{The complete structure of the proposed CNN-Transformer Cooperation Network (CTCNet). CTCNet is a U-shaped symmetrical hierarchical network with three stages: encoding stag, bottleneck stage, and decoding stage. Among them, the encoding stage is designed to extract local and global features with different scales, and the decoding stage is designed for feature fusion and image reconstruction.}
	\label{Network structure}
\end{figure*}

\begin{itemize}
\item We devise an efficient Feature Enhancement Unit (FEU) to reduce the feature redundancy of the network. Based on FEU, We further propose the Feature Refinement Module (FRM) to strengthen the different face structure information and enhance the extracted features.
\item We propose an efficient Local-Global Feature Cooperation Module (LGCM), which is composed of the carefully devised Facial Structure Attention Unit (FSAU) and a Transformer block, which can simultaneously capture local facial texture details and global facial structures for high-quality face super-resolution image reconstruction. 
\item We propose an elaborately designed Multi-scale Feature Fusion Unit (MFFU) to fuse the dense features from different scales and depths of the network. This operation ensures that our model can obtain rich features to better reconstruct high-quality images.
\item We devise a novel CNN-Transformer Cooperation Network (CTCNet) for face super-resolution based on LGCM and MFFU, which gains state-of-the-art performance in terms of various kinds of metrics.
\end{itemize}

\section{Related Work}
\label{sec2}

\subsection{Face Super-Resolution}
\label{sec21}

Due to the powerful feature representation capabilities of deep convolution neural networks (CNNs), significant progress has been made in nature image super-resolution~\cite{wang2020deep,li2021beginner,gao2022lightweight}. Li et al.~\cite{MSRN} designed the novel multi-scale residual network to fully interact and exploit the image features from different scales to enhance information. Guo et al.~\cite{DRN} presented a closed-loop dual regression network (DRN), which introduced an additional constraint to limit the mapping space between high- and low-resolution images. Zhang et al.~\cite{GLADSR} presented a global and local adjustment network to enhance the network capacity. Gao et al.~\cite{gao2022feature} designed a feature distillation interaction weighting network by making full use of the intermediate layer features.

CNN-based super-resolution methods have also greatly promoted the progress of face super-resolution (FSR). For example, 
Zhang et al.~\cite{zhang2018super} proposed a super-identity CNN, which introduced super-identity loss to assist the network in generating super-resolution face images with more accurate identity information. 
Lu et al.~\cite{lu2021face} devised a split-attention in split-attention network based on their designed external-internal split attention group for clear facial image reconstruction. In addition, some scholars have considered the particularity of the FSR task and proposed some FSR models guided by facial priors (e.g., face parsing maps and landmarks). 
Chen et al.~\cite{chen2018fsrnet} proposed the first end-to-end face super-resolution convolution network, which utilized facial parsing maps and landmark heatmaps to guide the super-resolution process. Kim et al.~\cite{kim2019progressive} also used face key point maps and face heatmaps to construct facial attention loss and used them to train a progressive generator. To tackle face images that exhibit large pose variations, Hu et al.~\cite{hu2020face} introduced the 3D facial priors to better capture the sharp facial structures. Ma et al.~\cite{ma2020deep} designed an iterative collaboration method that focuses on facial recovery and landmark estimation respectively. Li et al.~\cite{li2020learning} incorporated face attributes and face boundaries in a successive manner together with self-attentive structure enhancement to super-resolved tiny LR face images. Although these models have achieved promising results, they require additional marking on the dataset, and the accuracy of priors will greatly affect the accuracy of the reconstruction results.

\subsection{Attention Mechanism}
\label{sec22}
In the past few decades, the attention mechanism has made prominent breakthroughs in various visual image understanding tasks, such as image classification\cite{hu2018squeeze,wang2020eca}, image restoration~\cite{zhang2018image,dai2019second,niu2020single,chen2020learning}, etc. The attention mechanism can give more attention to key features, which benefits feature learning and model training. Zhang et al.~\cite{zhang2018image} proved that by considering the interdependence between channels and adjusting the channel attention mechanism, high-quality images could be reconstructed. 
Chen et al.~\cite{chen2020learning} presented a facial spatial attention mechanism, which uses the hourglass structure to form an attention mechanism. Therefore, the convolutional layers can adaptively extract local features related to critical facial structures.

Recently, Transformer~\cite{vaswani2017attention,devlin2018bert} are also widely used in computer vision tasks, such as image recognition~\cite{dosovitskiy2020image,touvron2021training}, object detection~\cite{carion2020end,zhu2020deformable}, and image restoration~\cite{liang2021swinir,lu2021efficient,esser2021taming,wang2021uformer,zamir2021restormer}. The key idea of the Transformer is the self-attention mechanism that can capture the long-range correlation between words/pixels. Although pure Transformers have great advantages in distilling the global representation of images, only depending on image-level self-attention will still cause the loss of local fine-grained details. Therefore, how effectively combining the global information and local features of the image is important for high-quality image reconstruction, which is also the goal of this work.


\section{CNN-Transformer Cooperation Network}
\label{sec3}

In this section, we first depict the overall architecture of the proposed CNN-Transformer Cooperation Network (CTCNet). Then, we introduce each module in the network in detail. Finally, we introduce related loss functions for supervised CTCGAN training.

\subsection{Overview of CTCNet}
\label{sec31}

As shown in Fig.~\ref{Network structure}, the proposed CTCNet is a U-shaped symmetrical hierarchical network with three stages: encoding stag, bottleneck stage, and decoding stage. Among them, the encoding stage is designed to extract local and global features with different scales, and the decoding stage is designed for feature fusion and image reconstruction. Meanwhile, multi-scale connections are used between the encoding stage and the decoding stage to achieve sufficient feature aggregation. To better demonstrate the model, we define ${I_{LR}}$, ${I_{SR}}$, and ${I_{HR}}$ as the LR input image, the recovered SR image, and the ground-truth HR image, respectively. 

\subsubsection{Encoding Stage} As we mentioned above, the encoding stage is designed for feature extraction. Therefore, giving a degraded image ${I_{LR}}$ as the input, we first apply a ${3\times3}$ convolution layer to extract the shallow features. After that, the extracted features are passed through 3 encoding stages. Each encoding stage includes one specially designed Local-Global Feature Cooperation Module (LGCM) and one downsampling block. Among them, LGCM consists of a Facial Structure Attention Unit (FSAU) and a Transformer block. The downsampling block consists of a ${3\times3}$ convolutional layer with stride 2, a LeakyReLU activation function, and a ${3\times3}$ convolution with stride 1, in which the first convolution uses stride 2 to extract feature information and reduce the size simultaneously. Therefore, after each encoding stage, the size of the output feature maps will be halved, while the number of output channels will be doubled. For instance, given the input feature maps ${I_{LR} \in \mathbb{R}^{C \times H \times W}}$, the $i$-th stage of the encoder produces the feature maps ${I_{en}^{i} \in \mathbb{R}^{2^{i}C \times \frac{H}{2^{i}} \times \frac{W}{2^{i}}}}$.

\subsubsection{Bottleneck Stage} There exists a bottleneck stage among the encoding and decoding stages. At this stage, all encoded features will be converged here. In order to make these features better utilized in the decode stage, we introduce Feature Refinement Module (FRM) to further refine and enhance the encoded features. With the help of FRMs, our model can focus on more facial structures and continuously strengthen different face structure information.

\begin{figure*}[t]
\centering
\includegraphics[width=18cm]{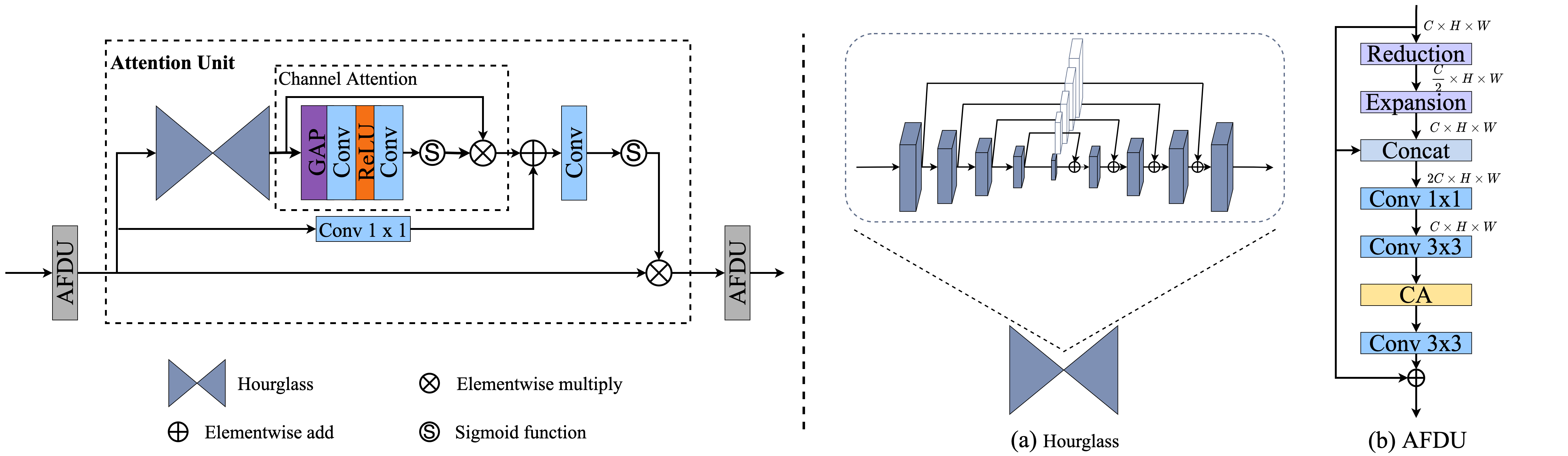}
\caption{The architecture of the proposed Facial Structure Attention Unit (FSAU). Among them, GPA is the Global Average Pooling operation.}
\label{FSAU}
\end{figure*}

\subsubsection{Decoding Stage} In the decoding stage, we focus on feature utilization and aim to reconstruct high-quality face images. To achieve this, we introduced a novel module, called Multi-scale Feature Fusion Unit (MFFU). Specifically, the decoder takes the latent features of the LR image as inputs and progressively fuses them through MFFUs to reconstruct the SR representations. As shown in Fig.~\ref{Network structure}, each decoder consists of an upsampling block, an MFFU, and an LGCM. Among them, the upsampling block consists of a ${6\times6}$ transposed convolutional layer with stride 2, a LeakyReLU activation function, and a ${3\times3}$ convolution with stride 1, in which the transposed convolutional layer uses stride 2 to extract feature information and increase the size of features simultaneously. Therefore, each decoder halves the number of the output feature channels while doubles the size of the output feature maps. It is worth mentioning that in MFFU, it will simultaneously fuses features with different scales extracted in the encoding stage. Therefore, all local and global features with different scales can be fully used to reconstruct high-quality face images. At the end of the decoding stage, we use a ${3\times3}$ convolutional layer to convert the learned features into the final SR features $I_{Out}$.


Finally, the high-quality SR face image is obtained by ${I_{SR}=I_{LR}+I_{Out}}$. Given a training dataset ${\left\{I_{LR}^{i}, I_{HR}^{i}\right\}_{i=1}^{N}}$, we optimize our \textit{\textbf{CTCNet}} by minimizing the following pixel-level loss function:
\begin{equation}
{\mathcal{L}(\Theta) = \frac{1}{N}\sum_{i=1}^{N}\left \|F_{CTCNet}(I_{LR}^{i},\Theta ) -I_{HR}^{i}\right\|_{1}},
\end{equation}
where $N$ denotes the number of training images. ${I_{LR}^{i}}$ and ${I_{HR}^{i}}$ are the LR image and the ground-truth HR image of the $i$-th image, respectively. Meanwhile, $F_{CTCNet}(\cdot)$ and $\Theta$ denote the CTCNet and its network parameters, respectively.

\subsection{Local-Global Feature Cooperation Module (LGCM)}

As one of the most important modules in CTCNet, LGCM is designed for local and global feature extraction. As shown in Fig.~\ref{Network structure}, LGCM consists of a Facial Structure Attention Unit (FSAU) and a Transformer Block, which are used for local and global feature extraction, respectively.

\subsubsection{Facial Structure Attention Unit (FSAU)}
\label{sec32}

In FSR, the main challenge is how to extract the key facial features (such as eyes, eyebrows, and mouth), and make the network pay more attention to these features. To achieve this, we propose the Facial Structure Attention Unit (FSAU) to make our model extract as much as possible useful information for better detail restoration. As shown in Fig.~\ref{FSAU}, FSAU mainly consists of one Attention Unit and two Adaptive Feature Distillation Units (AFDU). In the Attention Unit, we use channel attention nested in spatial attention to better extract spatial features and promote channel information interaction. This is because combining the two attention mechanisms can promote the representation power of the extracted features. Specifically, we first adopt the hourglass structure to capture facial landmark features at multiple scales since the hourglass structure has been successfully used in human pose estimation and FSR tasks~\cite{chen2017adversarial,newell2016stacked}. After that, in order to make the module focus on the features of the critical facial components, we introduce the channel attention (CA) mechanism~\cite{zhang2018image} to pay more attention to the channels containing landmark features. Then, we use an additional ${3\times3}$ convolutional layer and Sigmoid function to generate the spatial attention maps of the key components of the face. Finally, to alleviate the problem of vanishing gradients, we also add the residual connection between the input of the hourglass and the output of CA. 

In addition, we also introduce Adaptive Feature Distillation Units (AFDUs) at the beginning and end of the attention unit for local feature extraction. As shown in Fig.~\ref{FSAU} (b), to save memory and the number of parameters, we first use the Reduction operation to halve the number of the feature maps and then restore it by the Expansion operation. Among them, Reduction and Expansion operations are both composed of a ${3\times3}$ convolutional layer. Meanwhile, we apply the concatenation operation to aggregate the input of Reduction and the output of Expansion along the channel dimension, followed by a ${1\times1}$ convolutional layer and a ${3\times3}$ convolutional layer. The ${1\times1}$ convolution is used to fully utilize the hierarchical features, while the ${3\times3}$ convolution is dedicated to reducing the number of feature maps. After that, a CA module is employed to highlight the channels with higher activated values, and a ${3\times3}$ convolutional layer is used to refine the extracted features. Finally, the residual learning mechanism~\cite{he2016deep} is also introduced to learn the residual information from the input and stabilize the training. 

\begin{figure}[t]
\centering
\includegraphics[width=8cm, trim=0 0 50 0]{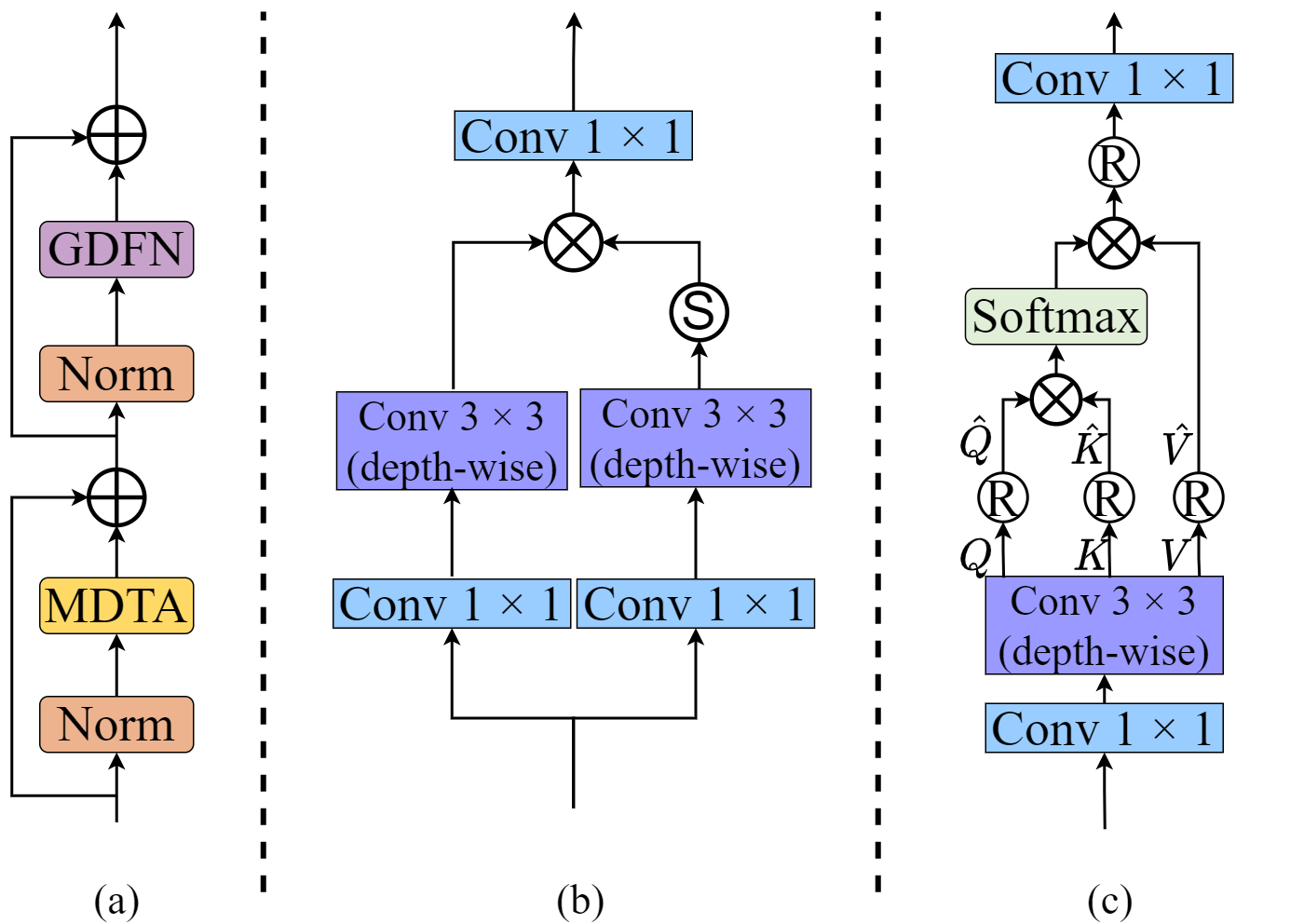}
\caption{The architecture of (a) Transformer Block (b) GDFN, and (c) MDTA, respectively.}
\label{Transformer Block}
\end{figure}

\subsubsection{Transformer Block}
\label{sec33}

As we mentioned above, FSAU is mainly designed for local feature extraction. However, this is far from enough to restore high-quality face images since the global facial structure (such as face contour) will be ignored due to the limited receptive field of CNN. To solve this problem, we introduce a Transformer Block to collaboratively learn the long-term dependence of images. 
Motivated by~\cite{zamir2021restormer}, in the multi-head self-attention part, we use the Multi-Dconv Head Transposed Attention (MDTA) to alleviate the time and memory complexity issues. 
Specifically, to make up for the limitations of the Transformer in capturing local dependencies, deep-wise convolution is introduced to enhance the local features to generate the global attention map. As depicted in Fig.~\ref{Transformer Block} (c), different from the original Transformer block directly achieved $query (Q)$, $key (K)$, and $value (V)$ by a linear layer, a ${1\times1}$ convolutional layer is used to aggregate pixel-level cross-channel context and a ${3\times3}$ depth convolutional layer is utilized to encode channel-level spatial context and generate ${Q,K,V\in \mathbb{R} ^{C\times H\times W}}$. Given the input feature $X\in \mathbb{R}^{C\times H\times W}$ and the layer normalized tensor $X^{'}\in \mathbb{R}^{C\times H\times W}$, we have
\begin{equation}
{Q = H_{pconv}^{1\times1}(H_{dconv}^{3\times3}(X^{'}))},
\end{equation}
\begin{equation}
{K = H_{pconv}^{1\times1}(H_{dconv}^{3\times3}(X^{'}))},
\end{equation}
\begin{equation}
{V = H_{pconv}^{1\times1}(H_{dconv}^{3\times3}(X^{'}))},
\end{equation}
where ${H_{pconv}^{1\times1}}(\cdot)$ is the ${1\times1}$ point-wise convolutional layer and ${H_{dconv}^{3\times3}}(\cdot)$ is the ${3\times3}$ depth-wise convolutional layer.

By calculating the correlation between $Q$ and $K$, we can obtain global attention weights from different locations, thereby capturing the global information. Next, we reshape $Q$, $K$, and $V$ into ${\hat{Q} \in \mathbb{R} ^{C\times HW}}$, ${\hat{K} \in \mathbb{R} ^{HW\times C}}$, and ${\hat{V} \in \mathbb{R} ^{C\times HW}}$, respectively. Thus the dot-product interaction of ${\hat{Q}}$ and ${\hat{K}}$ will generate a transposed-attention map with size ${\mathbb{R}^{C\times C}}$, rather than the huge size of ${\mathbb{R}^{HW\times HW}}$. After that, the global attention weights are subsequently multiplied with $V$ to get the weighted integrated features $X_{w}\in \mathbb{R}^{C\times HW}$. This can help the module to capture valuable local context. Finally, we reshape $X_{w}$ into $\hat{X_{w}}\in \mathbb{R}^{C\times H\times W}$ and use a ${1\times1}$ convolutional layer to realize feature communication. The above procedure can be formulated as follows:
\begin{equation}
{X_{weighted} = \operatorname{Softmax}(\hat{Q}\cdot\hat{K}/\sqrt{d})} \cdot \hat{V},
\end{equation}
\begin{equation}
{Y_{M} = H_{pconv}^{1\times 1}(R(X_{weighted}))},
\end{equation}
where ${Y_{M}}$ denotes the output of MDTA, ${R(\cdot)}$ stands for the reshaping operation. Here, ${\sqrt{d}}$ is a temperature parameter to control the magnitude of the dot product of ${\hat{K}}$ and ${\hat{Q}}$ before applying the Softmax function. 

\begin{figure}[t]
\centering
\includegraphics[width=9.3cm]{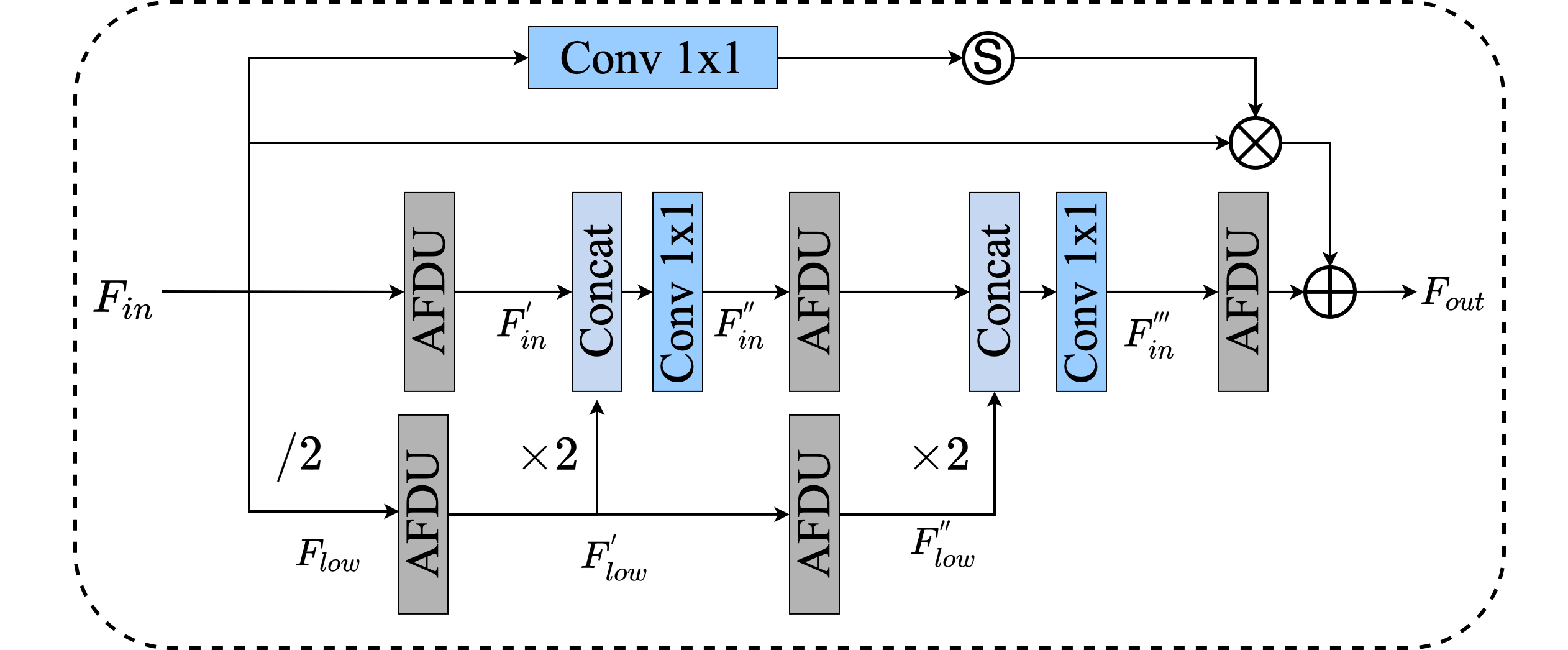}
\caption{The architecture of proposed FEU}
\label{FEU}
\end{figure}

\begin{figure*}[t]	\centerline{\includegraphics[width=16cm]{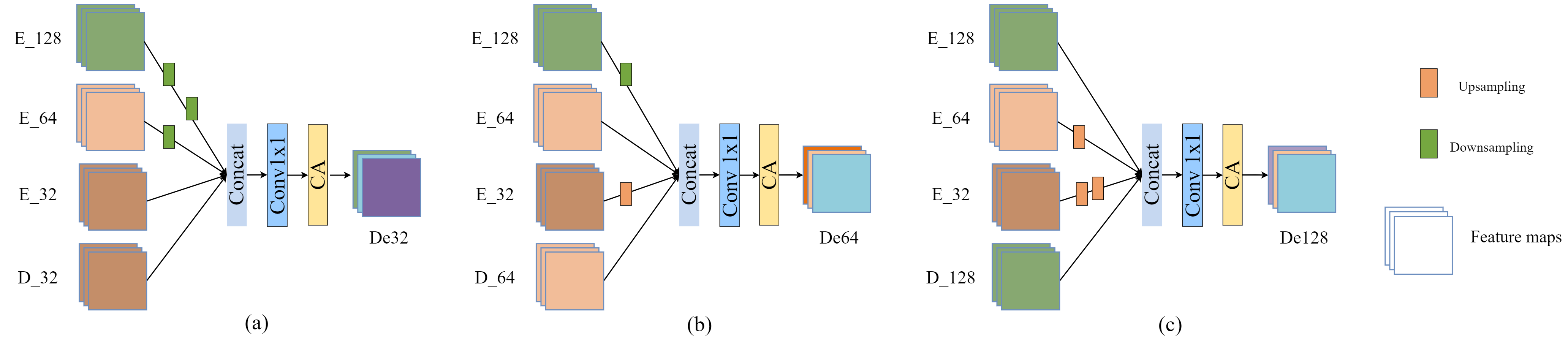}}
	\caption{Schematic diagram of how Multi-scale Feature Fusion Unit (MFFU) aggregates features from different scales.}
	\label{MFFU}
\end{figure*}

At the same time, we also introduce depth-wise convolutions into Gated-Dconv Feed-Forward Network (GDFN) to encode information from spatially neighboring pixel positions, responsible for learning local image structures for effective restoration. Given the input ${x}$, we have 
\begin{equation}\
{x^{'} = H_{dconv}^{3\times 3} (H_{pconv}^{1\times 1}(x))},
\end{equation}
\begin{equation}\
{Y_{G}  = H_{pconv}^{1\times 1} (x^{'} \cdot \sigma (x^{'}))},
\end{equation}
where ${\sigma}$ denotes the GELU non-linearity operation~\cite{Dan2016gauss} and $Y_{G}$ denotes the output of GDFN.

With the help of FSAU and Transformer Block, LGCM is able to capture both local features and global relationships of faces, which is beneficial for high-quality image reconstruction.

\subsection{Feature Refinement Module (FRM)}
\label{sec34}

In the bottleneck stage, we introduce the well-designed Feature Refinement Modules (FRMs) to continuously refine and enhance the important encoded features of the face. As shown in Fig.~\ref{Network structure}, each FRM encompasses an FSAU and a Feature Enhancement Unit (FEU). To reduce the computational burden and feature redundancy of the network, we use a double-branch structure in FEU. As shown in Fig.~\ref{FEU}, the first branch mainly uses AFDUs to extract the information in the original scale, while the second branch extracts features from the down-sampled feature maps, which are then up-sampled to fuse with the outputs of the first branch. In comparison with the general residual learning, we also add a feature self-calibration path to the residual connection to fully mine the hierarchical features and stabilize the training simultaneously. 
The above operations can be expressed as
\begin{equation}
{F_{in}^{\prime}=f_{a}\left(F_{in}\right),F_{low}^{\prime}=f_{a}\left(\downarrow F_{in}\right)},F_{low}^{\prime \prime}=f_{a}(F_{low}^{\prime}),
\end{equation}
\begin{equation}
{F_{in}^{\prime\prime}=H_{conv}^{1\times1}\left(H_{cat}\left(f_{a}\left(F_{in}^{\prime}\right),\uparrow f_{a}\left(F_{low}^{\prime}\right)\right)\right.},
\end{equation}
\begin{equation}
{F_{in}^{\prime\prime\prime}=H_{con v}^{1\times1}\left(H_{cat}\left(f_{a}\left(F_{in}^{\prime\prime}\right), \uparrow f_{a}\left(F_{l o w}^{\prime \prime}\right)\right)\right.},
\end{equation}
\begin{equation}
F_{out}=f_{a}\left(F_{in}^{\prime \prime \prime}\right)+F_{in} \cdot \sigma\left({H}_{conv}^{1 \times 1}\left({~F}_{in}\right)\right),
\end{equation}
where ${f_{a}(\cdot)}$ denotes the operation of AFDU, ${H_{cat}(\cdot)}$ indicates the feature concatenating operation along the channel dimension, ${H_{conv}^{1 \times 1}(\cdot)}$ stands for the ${1\times1}$ convolutional layer, and ${\sigma}$ denotes the Sigmoid function.

\subsection{Multi-scale Feature Fusion Unit (MFFU)}
\label{sec35}

In order to make full use of the multi-scale features extracted in the encoding stage, we introduce the multi-scale feature fusion scheme in the decoding stage to enable the network to have better feature propagation and representation capabilities. Specifically, our main goal is to explore and exploit the features from the encoding stage during the decoding process. However, the sizes of these features are different, and how to integrate these features more effectively is critically important. Take the size of the input image as ${128\times128}$ as an example, the size of the feature maps we obtained in the encoding stages is ${128\times128}$, ${64\times64}$, and ${32\times32}$, respectively. However, the size of the feature maps in the decoding stage is ${32\times32}$, ${64\times64}$, and ${128\times128}$, successively. To solve this problem, we design a Multi-scale Feature Fusion Unit (MFFU). The details of MFFU are given in Fig~\ref{MFFU}. According to the figure, we can observe that we first use upsampling and downsampling operations to scale the image feature maps with inconsistent sizes. After unifying the size of all feature maps, we concatenate the four types of feature maps along the channel dimension. Then, we use a ${1\times1}$ convolutional layer to generate the preliminary fusion result. Finally, we assign a channel direction attention weight to each channel through the CA mechanism.

Based on the size of the feature maps, the fusion scheme can be divided into three situations. The schematic diagram of how MFFU aggregates features from different scales is shown in Fig~\ref{MFFU}. For the sake of simplicity, we only give the formulation of Fig~\ref{MFFU} (b). 
The formulation of Fig {\ref{MFFU} (b)} can be defined as:
\begin{equation}
{E_{128_{-}64}=H_{conv}^{k3s2}\left(E_{128}\right)},
\end{equation}
\begin{equation}
{E_{32_{-}64}=H_{deconv}^{k6s2p2}\left(E_{32}\right)},
\end{equation}
\begin{equation}
De_{64}^{\prime}=H_{{conv}}^{k1s 1}\left(H_{cat}\left(E_{128\_64}, E_{32\_64}, E_{64}, D_{64}\right)\right),
\end{equation}
\begin{equation}
De_{64}=CA\left(De_{64}^{\prime}\right),
\end{equation}
where ${E_{k}(k=32,64,128)}$ represents the feature maps from the previous three encoding stages with the size of ${k\times k}$, and ${D_{64}}$ represents the original feature maps of the current decoder with the size of ${64\times 64}$. ${E_{m\_n}}$ indicates that the size of the feature maps has changed from ${m\times m}$ to ${n\times n}$. ${H_{conv}^{k3s2}(\cdot)}$ denotes the ${3\times3}$ convolution operation with the stride to be 2, while ${H_{deconv}^{k6s2p2}(\cdot)}$ denotes the ${6\times6}$ transposed convolution operation with stride and padding to be 2. ${H_{cat}(\cdot)}$ denotes the concatenating operation along the channel dimension. ${De_{64}^{'}}$ represents the preliminary fusion result and ${De_{64}}$ means the final fusion result.

\subsection{Model Extension}
\label{sec36}

As we know, Generative Adversarial Network (GAN) has been proven to be effective in recovering photo-realistic images~\cite{ledig2017photo,wang2018esrgan}. Therefore, we also extended our model with GAN and propose an extended model in this work, named CNN-Transformer Cooperation Generative Adversarial Network (CTCGAN). In CTCGAN, we use our CTCNet as the generative model and utilize the discriminative model in the conditional manner~\cite{isola2017image}. The new loss functions adopted in training the CTCGAN consist of three parts:

\subsubsection{Pixel Loss} The same as CTCNet, we use pixel-level loss to constrain the low-level information between the SR image and the HR image. It is can be defined as
\begin{equation}
{\mathcal{L}_{pix}=\frac{1}{N} \sum_{i=1}^{N}\left\|G(I_{LR}^{i})-I_{HR}^{i}\right\|_{1}},
\end{equation}
where $G(\cdot)$ indicates the CTCGAN generator.

\subsubsection{Perceptual Loss} The perceptual loss is mainly used to promote the perceptual quality of the reconstructed SR images. Specifically, we use a pre-trained face recognition VGG19~\cite{simonyan2014very} to extract the facial features. Therefore, we can calculate the feature-level similarity of the two images. The perceptual loss can be defined as
\begin{equation}
{\mathcal{L}_{pcp}=\frac{1}{N}\sum_{i=1}^{N}\sum_{l=1}^{L_{VGG}}\frac{1}{M_{VGG}^{l}}\left\|f_{VGG}^{l}\left(I_{SR}^{i}\right)-f_{VGG}^{l}\left(I_{HR}^{i}\right)\right\|_{1}}, 
\end{equation}
where $f_{VGG}^{l}(\cdot)$ is the $l$-th layer in $VGG$, $L_{VGG}$ denotes the total number of layers in $VGG$, and $M_{VGG}^{l}$ indicates the number of elements in $f_{VGG}^{l}$.

\subsubsection{Adversarial Loss} The principle of GAN is that generator $G$ strives to create fake images, while discriminator $D$ tries to distinguish fake pictures. In other words, the discriminator $D$ aims to distinguish the super-resolved SR image and the HR image by minimizing
\begin{equation}
{\mathcal{L}_{dis}=-\mathbb{E}\left[\log\left(D\left(I_{H R}\right)\right)\right]-\mathbb{E}\left[\log\left(1-D\left(G\left(I_{L R}\right)\right)\right)\right]}.
\end{equation}


\begin{table}[!t]
	\begin{center}
		\caption{Verify the effectiveness of LGCM on CelebA ($\times 8$).}
		\setlength{\tabcolsep}{3mm}
        \renewcommand\arraystretch{1}
		\begin{tabular}{c|cccc}
			\hline
			Methods      & PSNR${\uparrow}$  & SSIM${\uparrow}$  & VIF${\uparrow}$  & LPIPS${\downarrow}$      \\
			\cline{2-5}
			\hline
			\hline
			w/o LGCM         & 27.56      &0.7867       & 0.4487       &0.2051 \\
			LGCM w/o TB      & 27.82      &0.7964       &0.4707        &0.1833 \\
			LGCM w/o FSAU    & 27.83      &0.7972       & 0.4637       &0.1845  \\
			LGCM             &\bf{27.90}  &\bf{0.7980}  &\bf{0.4721}   &\bf{0.1797}  \\
			\hline
		\end{tabular}
		\label{Effects of LGCM}
	\end{center}
\end{table}

\begin{table}[!t]
 \begin{center}
  \caption{Performance comparisons of different numbers of FRM on CelebA ($\times 8$).}
  \setlength{\tabcolsep}{2.8mm}
        \renewcommand\arraystretch{1}
  \begin{tabular}{c|cccc}
   \hline
   Methods      &PSNR/SSIM${\uparrow}$   &VIF${\uparrow}$  &LPIPS${\downarrow}$   &Parameters${\downarrow}$     \\
   \cline{2-5}
   \hline
   \hline
   CTCNet-V0     &27.77/0.7954        &0.4683       &0.1856       &\bf{10.416M} \\
            CTCNet-V2     &27.83/0.7965        &0.4692       &0.1858       &16.014M      \\
            CTCNet-V4     &\bf{27.87/0.7979}   &\bf{0.4728}  &\bf{0.1834}  &21.613M      \\ 
            CTCNet-V6     &27.85/0.7967        &0.4691       &0.1872       &27.212M      \\
   \hline
  \end{tabular}
 \label{Numbers of LFRM}
 \end{center}
\end{table}

\begin{table}[!t]
 \begin{center}
  \caption{Performance comparisons of different feature fusion methods in MFFU. The last line is the strategy used in our final model (CelebA, $\times 8$).}
  \setlength{\tabcolsep}{4.1mm}
        \renewcommand\arraystretch{1}
  \begin{tabular}{cccc|cccc}
  \hline
        MSC   &Concat  &Add  &CA &PSNR${\uparrow}$  &SSIM${\uparrow}$ \\
        \hline
        \hline
        ${\times}$  & ${\times}$  & ${\times}$  & ${\times}$  & 27.76 &0.7961  \\ 
        $\surd$     & $\surd$     & ${\times}$  & ${\times}$  & 27.84 &0.7969  \\ 
        $\surd$     & ${\times}$  & $\surd$     & ${\times}$  & 27.82 &0.7955 \\  
        $\surd$     & ${\times}$  & $\surd$     & $\surd$     & 27.83 &0.7960 \\  
        $\surd$     & $\surd$     & ${\times}$  & $\surd$     & \bf{27.87}  &\bf{0.7979}\\   
        \hline
  \end{tabular}
  \label{Effects of MFFU}
 \end{center}
\end{table}

In addition, the generator tries to minimize 
\begin{equation}
\mathcal{L}_{adv}=-\mathbb{E}\left[\log \left(D\left(G\left(I_{L R}\right)\right)\right)\right].
\end{equation}

 Therefore, \textit{\textbf{CTCGAN}} is optimized by minimizing the following overall objective function:
 
\begin{equation}
{\mathcal{L}=\lambda _{pix}\mathcal{L}_{pix}+\lambda_{pcp}\mathcal{L}_{pcp}+\lambda_{adv}\mathcal{L}_{adv}},
\end{equation}
where $\lambda _{pix}$, $\lambda_{pcp}$, and $\lambda_{adv}$ indicate the trade-off parameters for the pixel loss, the perceptual loss, and the adversarial loss, respectively.

\section{Experiments}
\label{sec4}

\subsection{Datasets}
\label{sec41}

In our experiments, we use CelebA~\cite{liu2015deep} dataset for training and evaluate the model validity on Helen~\cite{le2012interactive} and SCface~\cite{grgic2011scface} datasets. The height and width of the face pictures in CelebA are inconsistent. Therefore, we crop the image according to the center point, and the size is adjusted to ${128\times128}$ pixels, which is used as the HR image. Then we down-sample these HR images into ${16\times16}$ pixels with the bicubic operation and treat them as the LR inputs. We use 18,000 samples of the CelebA dataset for training, 200 samples for validating, and 1,000 samples for testing. Furthermore, we also directly test our model on Helen and SCface datasets using the model trained on CelebA.

\subsection{Implementation Details}
\label{sec42}

We implement our model using the PyTorch framework. Meanwhile, we optimize our model by Adam and set ${\beta _{1} = 0.9}$ and ${\beta _{2} = 0.99}$. The initial learning rate is set to ${2\times 10^{-4}}$. For CTCGAN, we empirically set ${\lambda _{pix} =1}$, ${\lambda _{pcp} =0.01}$, and ${\lambda _{adv} =0.01}$. We also use Adam to optimize both $G$ and $D$ with ${\beta _{1} = 0.9}$ and ${\beta _{2} = 0.99}$. The learning rates of $G$ and $D$ are set to ${1\times 10^{-4}}$ and ${4\times 10^{-4}}$, respectively. 

To assess the quality of the SR results, we employ four objective image quality assessment metrics: Peak Signal to Noise Ratio (PSNR), Structural Similarity (SSIM)~\cite{wang2004image}, Learned Perceptual Image Patch Similarity (LPIPS)~\cite{zhang2018unreasonable}, and Visual Information Fidelity (VIF)~\cite{sheikh2006image}.

\begin{figure}[t]
\centering
\includegraphics[width=0.99\columnwidth]{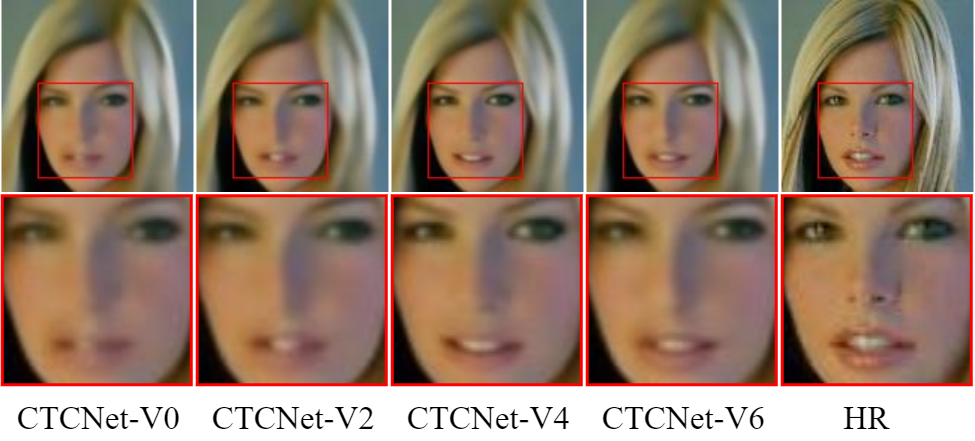}
\caption{Visual comparisons of different numbers of FRM on CelebA dataset for $\times 8$ SR.}
\label{fNumbers of LFRM}
\end{figure}

\begin{figure}[t]
\centering
\includegraphics[width=0.99\columnwidth]{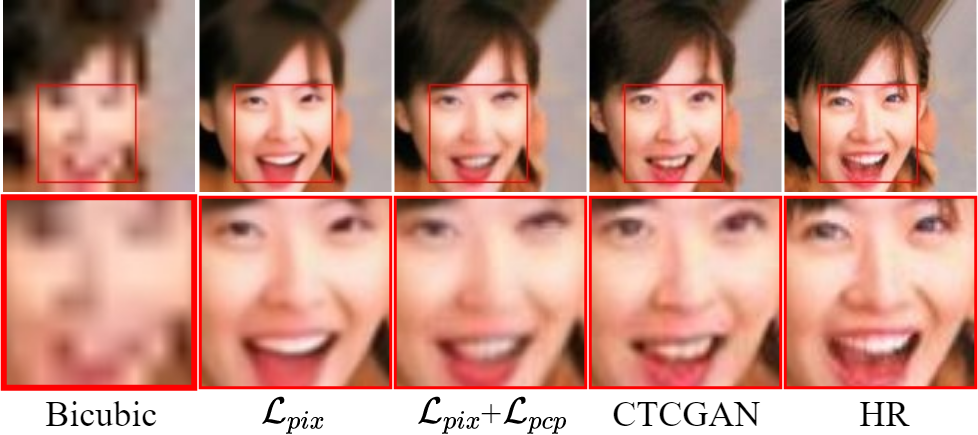}
\caption{Ablation study of losses used in CTCGAN on CelebA dataset for $\times 8$ SR.}
\label{fEffects of GAN_loss}
\end{figure}

\subsection{Ablation Studies}
\label{sec43}
In this part, we provide a series of ablation studies to verify the effectiveness of the model. Meanwhile, all ablation studies are conducted on the CelebA test sets for $\times$8 SR.

\subsubsection{Effectiveness of LGCM} LGCM is the most important module in CTCNet, which is designed to extract local features and global relationships of the image. At the same time, this is a new attempt to combine CNN and Transformer structures. To verify the effectiveness of LGCM and the feasibility of this combined method, we carried out a series of ablation studies in this part. As we know, LGCM contains an FSAU and a Transformer Block (TB). Therefore, design three modified models. The first model removes all LGCMs in the encoding and decoding stages, marked as ``w/o LGCM''. The second model removes all FSAUs while retaining the Transformer Block, marked as ``LGCM w/o FSAU''. The third model removes all Transformer Blocks while retaining the FSAU in LGCM, marked as ``LGCM w/o TB''. In Table~\ref{Effects of LGCM}, we show the results of these modified networks. 
According to the table, we have the following observations: 
(a) By comparing the first and the last lines in Table~\ref{Effects of LGCM}, we can observe that the introduced LGCM can significantly improve the performance of the model. This fully verifies the effectiveness of LGCM; (b) By comparing the first three lines, we can see that the performance of the model can also be improved by introducing FSAU or TB alone. This is because both local features and global relationships of the image are helpful for image reconstruction; (c) By comparing the last three lines, we can clearly observe that both FASU and TB play a unique role in FSR tasks. This is because FSAU can capture the local details while TB can capture the global facial structures simultaneously, which provides complementary information for the final SR image reconstruction. Only using one of them cannot achieve the best results. This further verifies the effectiveness of LGCM and the feasibility of combining CNN with Transformer.

\subsubsection{Effectiveness of FRM} To evaluate the effectiveness of FRM, we change the number of FRM in the bottleneck stage. We gradually increase the numbers of FRMs and denote the model with $N$ FRMs as CTCNet-VN, where ${N\in\left\{ 0,2,4,6\right\}}$. From Table~\ref{Numbers of LFRM}, we can observe that the model achieves the worst results when all FRMs are removed (CTCNet-V0). This illustrates the necessity of the existence of FRM in CTCNet. Meanwhile, it can be observed that the model performance can be improved with the increase of FRM within a certain range. However, we also notice that when the number of FRM exceeds $4$, the model performance will decrease and the model size will become larger. Therefore, we set $N$ = 4 to achieve a good balance between model performance and size. Meanwhile, from Fig.~\ref{fNumbers of LFRM}, we can intuitively see that as the number of FRM gradually increases from $0$ to $4$, the facial contours gradually become clear, which fully demonstrates the effectiveness of stacking multiple FRMs.

\begin{table}[!t]
 \begin{center}
  \caption{Study of each component in FSAU (CelebA, $\times 8$).}
  \setlength{\tabcolsep}{4mm}
        \renewcommand\arraystretch{1}
  \begin{tabular}{cc|cccc}
   \hline
   CA  &SA  &PSNR${\uparrow}$ &SSIM${\uparrow}$  &VIF${\uparrow}$  &LPIPS${\downarrow}$   \\
   \cline{2-5}
   \hline
   \hline
   ${\times}$   &${\times}$   &27.80  &0.7989  &0.4701  &0.1874               \\ 
            $\surd$      &${\times}$   &27.83  &0.7966  &0.4673  &0.1881               \\ 
            ${\times}$   &$\surd$      &27.82  &0.7964  &0.4676  &0.1908               \\ 
            $\surd$      &$\surd$      &\bf{27.87}  &\bf{0.7979}  &\bf{0.4728}  &\bf{0.1834}          \\   \hline 
  \end{tabular}
  \label{Effects of FSAU}
 \end{center}
\end{table}

\begin{table}[!t]
 \begin{center}
  \caption{Study of each component in FEU (CelebA, $\times 8$).}
  \setlength{\tabcolsep}{3.5mm}
        \renewcommand\arraystretch{1}
  \begin{tabular}{c|cccc}
   \hline
   Methods & PSNR${\uparrow}$  & SSIM${\uparrow}$  & VIF${\uparrow}$  & LPIPS${\downarrow}$      \\
   \cline{2-5}
   \hline
   \hline
   FEU w/o AFDU     & 27.77      &0.7947        & 0.4628       &0.1952      \\
   FEU w/o path     & 27.80      &0.7959        & 0.4659       &0.1907       \\
   FEU w/o dual     & 27.81      &0.7951        &0.4679        &0.1933      \\
   FEU              &\bf{27.87}  &\bf{0.7979}   &\bf{0.4728}   &\bf{0.1834}  \\
   \hline
  \end{tabular}
  \label{Effects of FEU}
 \end{center}
\end{table}

\subsubsection{Effectiveness of MFFU} MFFU is specially designed for multi-scale feature fusion. In this part, we conduct a series of experiments to demonstrate the effects of Multi-Scale Connections (MSC) and various feature fusion methods in MFFU. The first experiment is used to verify the necessity of MSC. The second and third experiments preserve the MSC but only use the concatenate or add operation to achieve multi-scale features fusion. The last two experiments use channel attention to reweigh the channels after the concatenate or add operation. From Table~\ref{Effects of MFFU}, it can be observed that (a) Using a multi-scale feature fusion strategy can effectively improve model performance, which proves the importance of multi-scale features for image reconstruction; (b) Using Channel Attention (CA) mechanism has positive effects on improving the model performance; (c) The effect of combining the concatenate operation and CA is apparent. This further verifies that adopting a suitable feature fusion strategy can well provide help for the subsequent reconstruction process. 

\begin{table}[!t]
 \begin{center}
  \caption{Verify the effectiveness of each loss component in CTCGAN (CelebA, $\times 8$).}
  \setlength{\tabcolsep}{2.5mm}
        \renewcommand\arraystretch{1}
  \begin{tabular}{c|cccc}
   \hline
   Methods & PSNR/SSIM${\uparrow}$  & VIF${\uparrow}$  & FID${\downarrow}$ & LPIPS${\downarrow}$      \\
   \cline{2-5}
   \hline
   \hline
   with $\mathcal{L}_{pix}$ (CTCNet)     & \bf{27.87/0.7979} &\bf{0.4728}   & 50.09       &0.1834       \\
   with $\mathcal{L}_{pix}$ and $\mathcal{L}_{pcp}$    & 27.43/0.7802      &0.4187        &30.83        &0.1694      \\
   CTCGAN              &27.38/0.7775       &0.4175        &\bf{30.64}   &\bf{0.1688}  \\
   \hline
  \end{tabular}
  \label{Effects of GAN_loss}
 \end{center}
\end{table}

\begin{table}[!t]
 \begin{center}
  \caption{Comparison results of GAN-based methods for $\times 8$ SR on the Helen test sets.}
  \setlength{\tabcolsep}{3.5mm}
        \renewcommand\arraystretch{1}
  \begin{tabular}{c|cccc}
   \hline
   Methods   & PSNR${\uparrow}$ & SSIM${\uparrow}$ &FID${\downarrow}$ &VIF${\uparrow}$ \\
   \cline{2-5}
   \hline
   \hline
   FSRGAN  &25.02   &0.7279 &146.55 &0.3400   \\
   DICGAN                          &25.59  &0.7398 &144.25 &0.3925 \\
   SPARNetHD                       &25.86  &0.7518 &149.54 &0.3932 \\
   CTCGAN (Ours)                    &\textbf{26.41} &\textbf{0.7776} & \textbf{118.05} & \textbf{0.4112} \\
   \hline
  \end{tabular}
 \label{Tab_GAN}
 \end{center}
\vspace{-0.2cm}
\end{table}

\subsubsection{Study of FSAU} In FSAU, we use the structure of the nested channel attention mechanism in the spatial attention mechanism to better extract spatial features and promote channel information interaction. To prove the effectiveness of using this nested structure, we remove channel attention and spatial attention respectively to perform ablation studies. From Table~\ref{Effects of FSAU}, we can see the effectiveness enlightened by the channel and spatial attention mechanisms. Adding channel attention or spatial attention alone can only slightly improve the PSNR value by 0.03dB and 0.02dB, respectively. However, when using the nested structure, the PSNR values increase from 27.80dB to 27.87dB. Therefore, we can draw a conclusion that we can gain better performance by applying the channel and spatial attention mechanisms simultaneously.

\subsubsection{Study of FEU} FEU is an essential part of FRM, which uses a double-branch structure to enhance feature extraction. As mentioned earlier, FEU mainly includes several AFDUs and a feature self-calibration path. 
In this part, we conducted three ablation experiments to verify the effectiveness of AFDU, dual-branch structure, and feature self-calibration path in FEU. From Table~\ref{Effects of FEU}, we can see that (a) If we do not use AFDU in FEU, the performance will drop sharply, and the usage of AFDU increases the PSNR value by 0.1dB; (b) Compared with a simple single-branch structure (without the downsampling and upsampling operations), using the dual-branch structure promotes the PSNR value by 0.06dB. It further verifies that multi-scale feature extraction often has better feature representation abilities; (c) The usage of the feature self-calculation path increases the PSNR value by 0.07dB, since this path can highlight the helpful features with higher activation values.

\begin{table*}[!t]
    \centering
    \caption{Quantitative comparisons for $\times$8 SR on the CelebA and Helen test sets.}
	\setlength{\tabcolsep}{3mm}
    \renewcommand\arraystretch{1}
	\scalebox{1}{
		\begin{tabular}{p{2cm}|p{1.3cm}p{1.3cm}p{1.3cm}p{1.3cm}|p{1.3cm}p{1.3cm}p{1.3cm}p{1.3cm}p{1.3cm}}
			\toprule
			\multirow{2}{*}{Methods}  & \multicolumn{4}{c|}{$CelebA$}   &\multicolumn{4}{c}{$Helen$}\\
			
			& PSRN${\uparrow}$ & SSIM${\uparrow}$ & VIF${\uparrow}$ & LPIPS${\downarrow}$ &PSNR${\uparrow}$  & SSIM${\uparrow}$ & VIF${\uparrow}$ & LPIPS${\downarrow}$  \\
			\hline
			\hline
	Bicubic                           & 23.61 & 0.6779 & 0.1821 & 0.4899 & 22.95 & 0.6762 & 0.1745 & 0.4912  \\
			
	SAN~\cite{dai2019second}          & 27.43 & 0.7826 & 0.4553 & 0.2080 & 25.46 & 0.7360 & 0.4029 & 0.3260  \\
			
	RCAN~\cite{zhang2018image}        & 27.45 & 0.7824 & 0.4618 & 0.2205 & 25.50 & 0.7383 & 0.4049 & 0.3437  \\
			
	HAN~\cite{niu2020single}          & 27.47 & 0.7838 & 0.4673 & 0.2087 & 25.40 & 0.7347 & 0.4074 & 0.3274  \\
			
	SwinIR~\cite{liang2021swinir}     & 27.88 & 0.7967 & 0.4590 & 0.2001 & 26.53 & 0.7856 & 0.4398 & 0.2644 \\
			
	FSRNet~\cite{chen2018fsrnet}      & 27.05 & 0.7714 & 0.3852 & 0.2127 & 25.45 & 0.7364 & 0.3482 & 0.3090  \\
			
	DICNet~\cite{ma2020deep}          & -     & -      & -      & -      & 26.15 & 0.7717 & 0.4085 & 0.2158 \\
			
	FACN~\cite{xin2020facial}         & 27.22 & 0.7802 & 0.4366 & 0.1828 & 25.06 & 0.7189 & 0.3702 & 0.3113  \\
			
	SPARNet~\cite{chen2020learning}   & 27.73 & 0.7949 & 0.4505 & 0.1995 & 26.43 & 0.7839 & 0.4262 & 0.2674  \\
			
	SISN~\cite{lu2021face}            & 27.91 & 0.7971 & 0.4785 & 0.2005 & 26.64 & 0.7908 & 0.4623 & 0.2571  \\
	    \midrule
	CTCNet (Ours)  &{\bf28.37} &{\bf0.8115} &{\bf0.4927} &{\bf0.1702} &{\bf27.08} &{\bf0.8077} &{\bf0.4732} &{\bf0.2094} \\
			
	    \bottomrule
		\end{tabular}
	}
	\label{compare_CelebA_Helen}
\end{table*}

\begin{figure*}[!t]
 \centerline{\includegraphics[width=18cm]{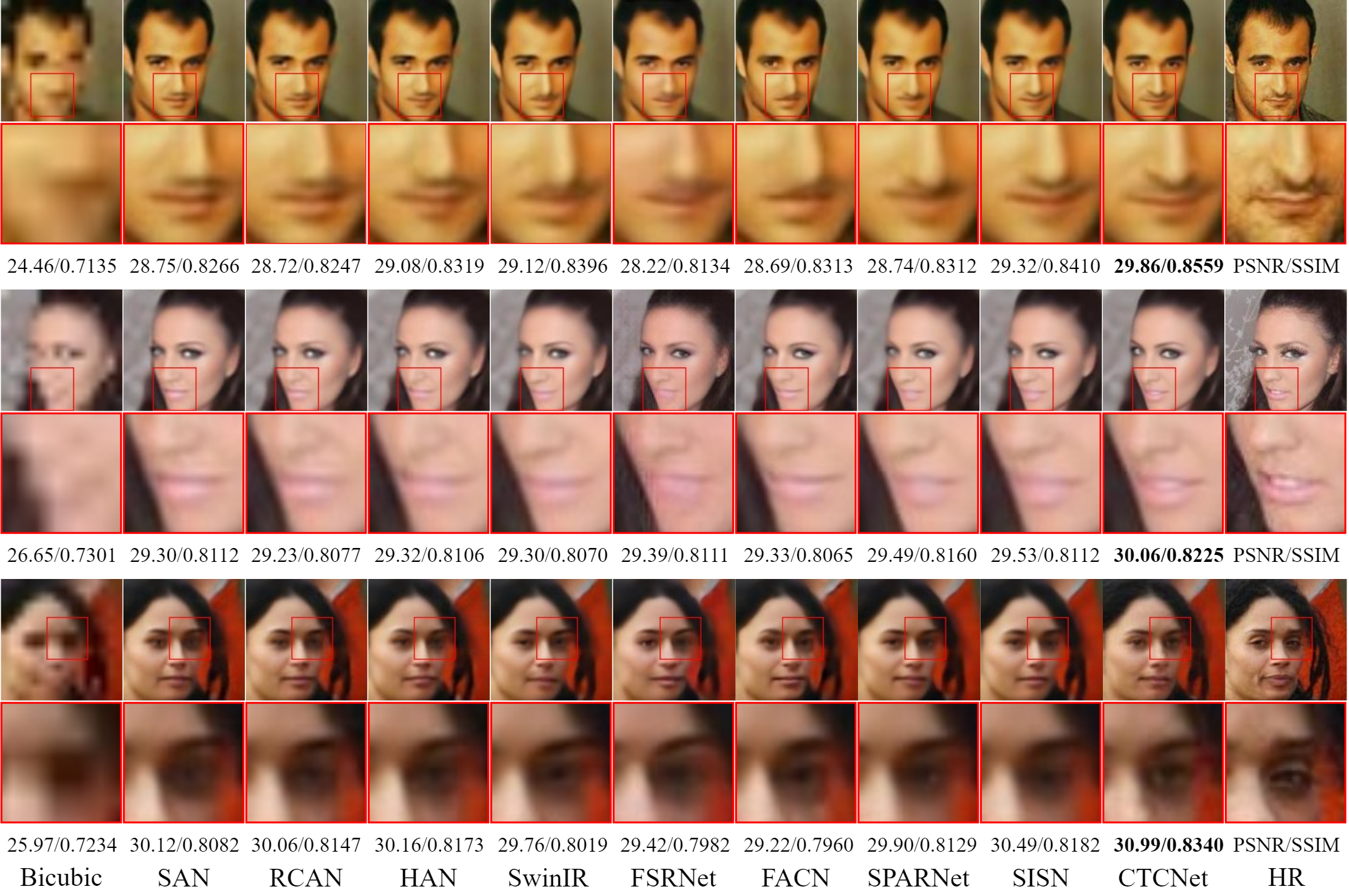}}
 \caption{Visual comparisons for $\times$8 SR on the CelebA test set. Obviously, our CTCNet can reconstruct clearer face images.}
 \label{compare_CelebA}
\end{figure*}

\subsubsection{Study of Loss Functions} To verify the effectiveness of different loss functions in CTCGAN, we conduct an ablation study by adding each of them progressively. The quantitative and qualitative comparisons are given in Table~\ref{Effects of GAN_loss} and Fig.~\ref{fEffects of GAN_loss}. We can observe that $\mathcal{L}_{pix}$ can produce better performance in terms of PSNR and SSIM, which are the generally used pixel-level-based image quality assessment metrics. From the 3-th and 4-th columns, we can see that the $\mathcal{L}_{pcp}$ and $\mathcal{L}_{adv}$ can generate photo-realistic images with superior visual effects than $\mathcal{L}_{pix}$. Although they can produce relatively sharp images, they tend to generate many false information and artifacts. 

\subsection{Comparison with Other Methods}
\label{sec44}

In this part, we compare our CTCNet with other state-of-the-art (SOTA) methods, including general image SR methods SAN~\cite{dai2019second}, RCAN~\cite{zhang2018image}, HAN~\cite{niu2020single}, novel FSR methods FSRNet~\cite{chen2018fsrnet}, DICNet~\cite{ma2020deep}, FACN~\cite{xin2020facial}, SPARNet~\cite{chen2020learning}, SISN~\cite{lu2021face}, and pioneer Transformer based image restoration method SwinIR~\cite{liang2021swinir}. For a fair comparison, all models are trained using the same CelebA dataset.

\begin{figure*}[t]
	\centerline{\includegraphics[width=18cm]{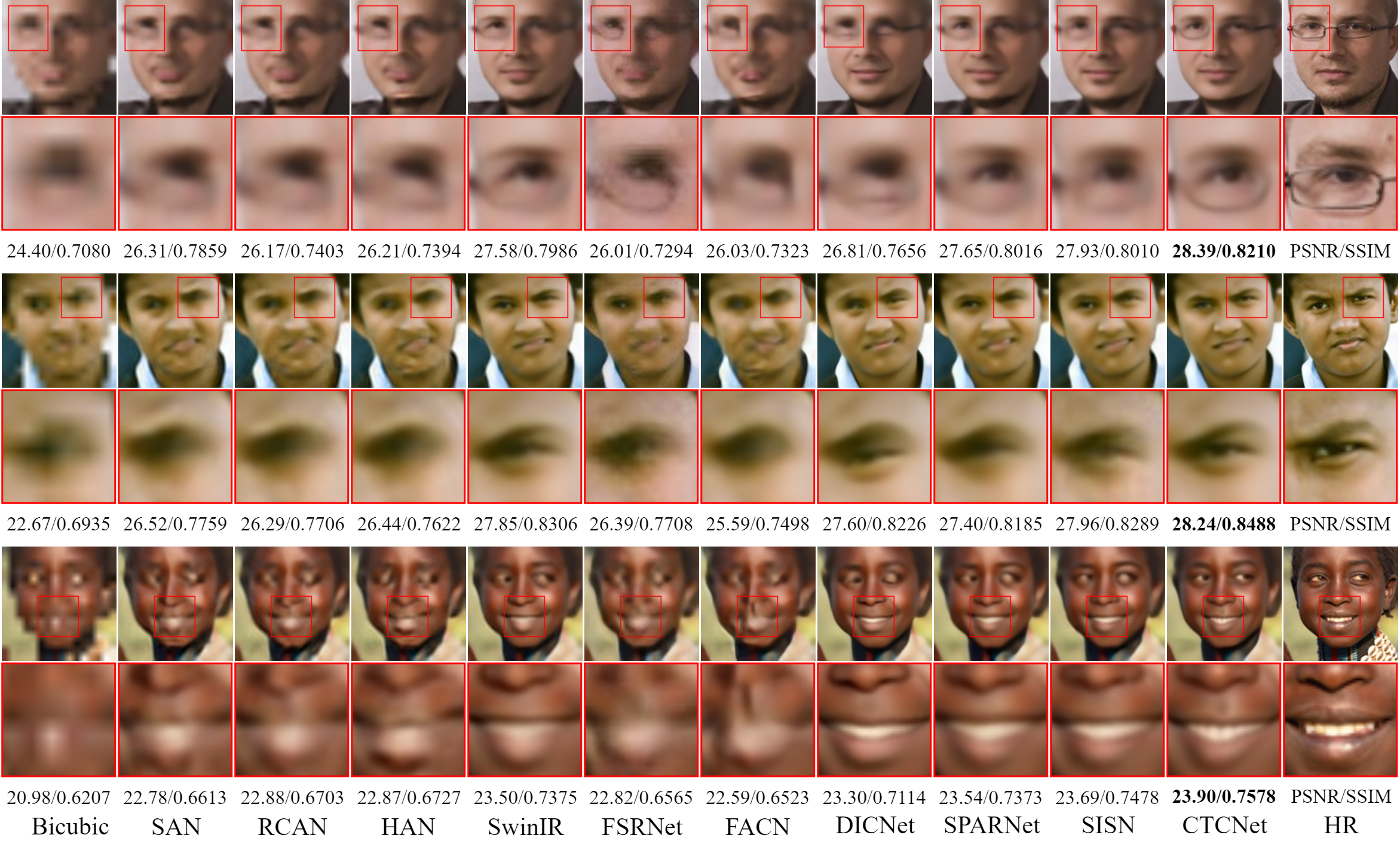}}
	\caption{Visual comparisons for $\times$8 SR on the Helen test set. Obviously, our CTCNet can reconstruct clearer face images.}
	\label{compare_Helen}
\end{figure*}

\begin{figure}[t]
\centering
\includegraphics[width=1\columnwidth]{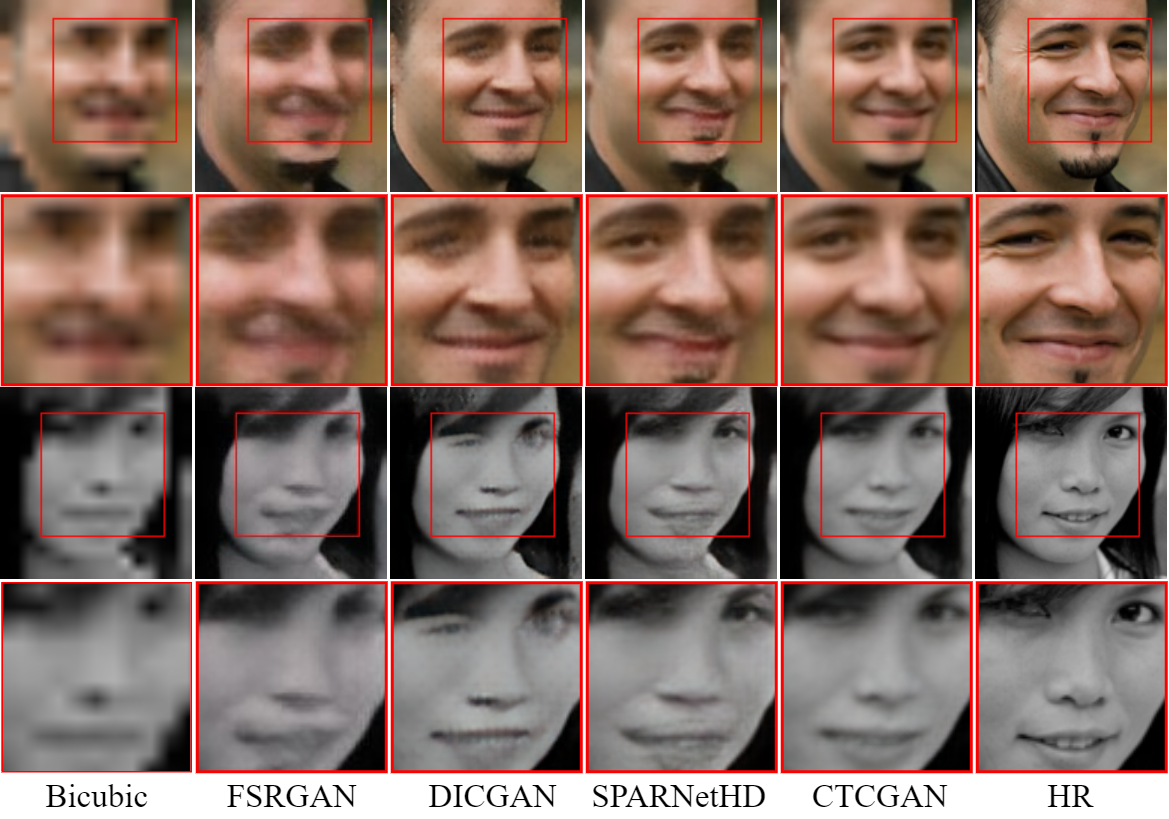}
\caption{Visual comparison of different GAN-based methods on the Helen test set. Obviously, our CTCGAN can reconstruct high-quality face images with clear facial components.}
\label{compare_GAN}
\end{figure}

\begin{figure}[t]
\centering
\includegraphics[width=1\columnwidth,trim=0 0 50 50]{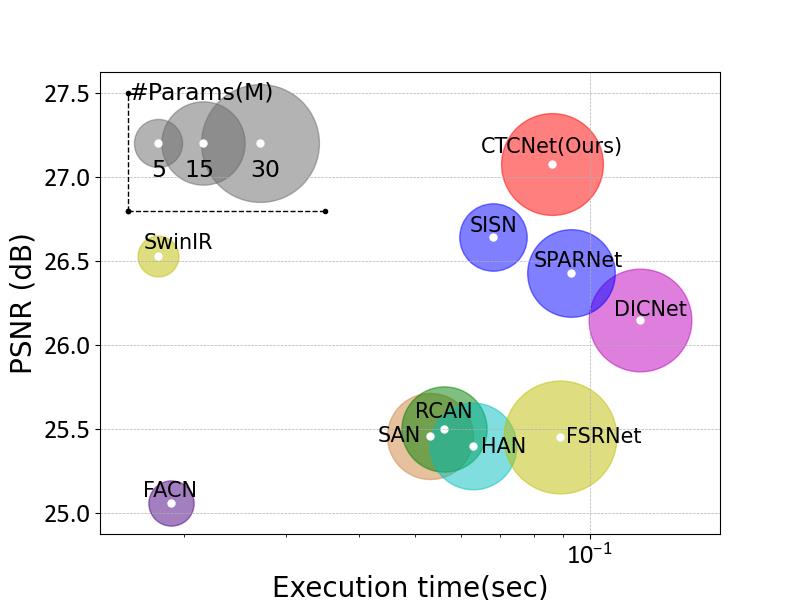}
\caption{Model complexity studies for $\times 8$ SR on the CelebA test sets. Our CTCNet achieves a better balance between model size, model performance, and execution time.}
\label{complexity_CelebA}
\end{figure}

\subsubsection{Comparison on CelebA dataset} The quantitative comparisons with other SOTA methods on the CelebA test set are provided in Table~\ref{compare_CelebA_Helen}. According to the table, we can see that CTCNet significantly outperforms other competitive methods in terms of PSNR, VIP, LPIPS, and SSIM. This fully verifies the effectiveness of CTCNet. Meanwhile, from the visual comparisons in Fig.~\ref{compare_CelebA} we can see that most of the previous methods cannot clearly restore the eyes and nose in the face, while our CTCNet can better restore face structures and generate more precise results. The reconstructed face images are closer to the real HR images, which further proves the effectiveness and excellence of CTCNet.

\begin{figure*}[!t]
\centering
\centerline{\includegraphics[width=17.8cm]{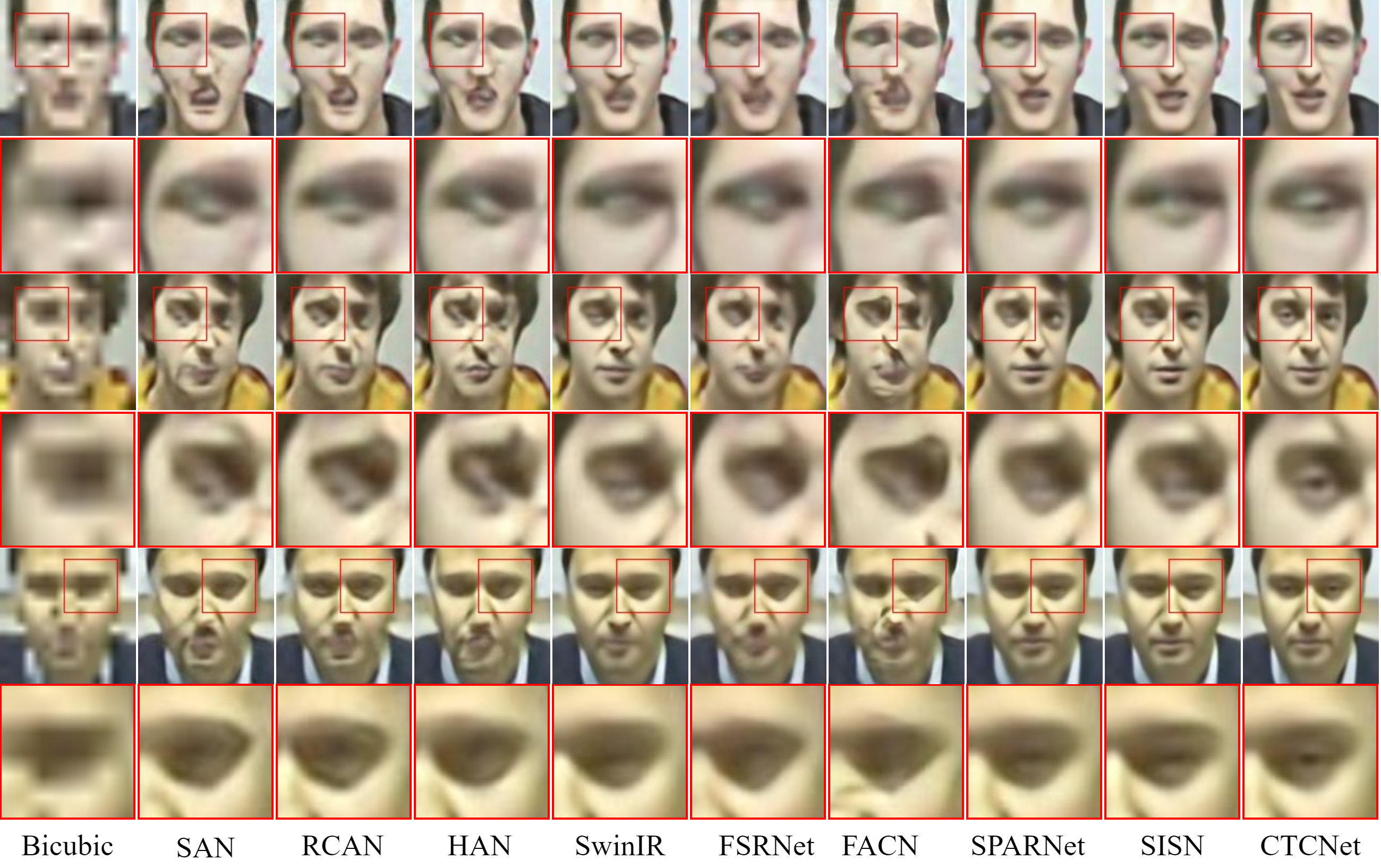}}
\caption{Visual comparison of respective methods on real-world surveillance scenarios for $\times$8 SR. Obviously, our CTCNet can reconstruct more clear and accurate details around the eyes and mouth.}
\label{compare_SCface}
\end{figure*}

\begin{table*}[!t] 
 \begin{center}
  \caption{Comparison results for average similarity of face images super-resolved by different methods.}
  \setlength{\tabcolsep}{2.5mm}
        \renewcommand\arraystretch{1}
  \scalebox{1}{
  \begin{tabular}{p{1.5cm}|p{1.1cm}|p{1.1cm}|p{1.1cm}|p{1.1cm}|p{1.1cm}|p{1.1cm}|p{1.1cm}|p{1.1cm}|p{1.1cm}|p{1.1cm}}
   \hline
   \multirow{2}{*}{Methods} &   \multicolumn{10}{c}{Average Similarity}     \\
   \cline{2-11}
   &Case 1 &Case 2 &Case 3 &Case 4 &Case 5 &Case 6 &Case 7 &Case 8 &Case 9 &Case 10\\
   \cline{2-11}
   \hline
   \hline
  SAN~\cite{dai2019second}        &0.8897 &0.9061 &0.9029 &0.8940 &0.8889 &0.9061 &0.9042 &0.8844 &0.9026 &0.9107 \\
  RCAN~\cite{zhang2018image}      &0.8927 &0.9000 &0.9038 &0.8957 &0.8963 &0.9090 &0.9028 &0.8807 &0.9045 &0.9064 \\
  HAN~\cite{niu2020single}        &0.8909 &0.9096 &0.8977 &0.9074 &0.8914 &0.9020 &0.9061 &0.8740 &0.8950 &0.9121 \\
  SwinIR~\cite{liang2021swinir}   &0.9087 &0.9196 &0.8991 &0.9079 &0.9105 &0.9040 &0.9119 &0.8939 &0.9080 &0.9093 \\
  FSRNet~\cite{chen2018fsrnet}    &0.8996 &0.8844 &0.9017 &0.8971 &0.8927 &0.9061 &0.8908 &0.8977 &0.9040 &0.9064 \\
  DICNet~\cite{ma2020deep}        &0.8859 &0.8814 &0.8692 &0.8760 &0.8736 &0.8755 &0.8837 &0.8743 &0.8687 &0.8914 \\
  FACN~\cite{xin2020facial}       &0.9048 &0.9009 &0.9040 &0.9017 &0.9058 &0.8985 &0.8970 &0.8906 &0.8687 &0.9007 \\
  SPARNet~\cite{chen2020learning} &0.9089 &0.9188 &0.8995 &0.9015 &0.9075 &0.8980 &0.9077 &0.9067 &0.9025 &0.9142 \\
  SISN~\cite{lu2021face}          &0.9127 &0.9206 &0.9086 &0.9049 &0.9080 &0.8999 &0.9175 &0.9098 &0.9060 &0.9227 \\
  \hline
  CTCNet     &\bf{0.9278}  &\bf{0.9219} &\bf{0.9129} &\bf{0.9165} &\bf{0.9243} &\bf{0.9194} &\bf{0.9228} &\bf{0.9136} &\bf{0.9106} &\bf{0.9280} \\
  \hline
  \end{tabular}}
 \label{Average_Similarity}
 \end{center}
\end{table*}

\subsubsection{Comparison on Helen dataset} In this part, we directly use the model trained on the CelebA dataset to test the model performance on the Helen test set to study the generality of CTCNet. Table~\ref{compare_CelebA_Helen} lists the quantitative experimental results on the Helen test set for $\times$8 SR. According to the table, we can clearly see that our CTCNet still achieves the best results on the Helen data set. From Fig.~\ref{compare_Helen}, we can also observe that the performance of most competitive methods degrades sharply, they cannot restore faithful facial details, and the shape is blurred. On the contrary, our CTCNet can still restore realistic facial contours and facial details. This further verifies the effectiveness and generality of CTCNet.

\subsubsection{Comparison with GAN-based methods} As we mentioned above, we also propose an extended model named CTCGAN. In this part, we compare our CTCGAN with three popular GAN-based FSR models: FSRGAN~\cite{chen2018fsrnet}, DICGAN~\cite{ma2020deep}, and SPARNetHD~\cite{chen2020learning}. As we all know, GAN-based SR methods usually have superior visual qualities but lower quantitative values (such as PSNR and SSIM). Therefore, we also introduce Frechet Inception Distance score (FID)~\cite{obukhov2020quality} as a new metric to evaluate the performance of GAN-based SR methods. In Table~\ref{Tab_GAN}, we provide the quantitative comparisons of these models on CelebA and Helen test sets. Obviously, our CTCGAN gains much better performance than other methods in terms of PSNR, SSIM, FID, and VIF. Meanwhile, the qualitative comparisons on the Helen test set are also provided in Fig.~\ref{compare_GAN}. According to the figure, we can see that those competitive methods cannot generate realistic faces and have undesirable artifacts and noise. In contrast, our CTCGAN can restore key facial components and texture details in the mouth and eyes. This fully demonstrates the effectiveness and excellence of our CTCGAN.

\subsubsection{Comparison on real-world surveillance faces} As we know, restoring face images from real-world surveillance scenarios is still a huge challenge. All the above experiments are in simulation cases, which can not simulate real-world scenarios well. To further verify the effectiveness of our CTCNet, we also conduct experiments on real-world low-quality face images, which are selected from the SCface dataset~\cite{grgic2011scface}. The images in SCface are captured by surveillance cameras, which inherently have lower resolutions hence no manual downsampling operation is required.

In this part, we try to restore the face images with more texture details and good facial structures. A visual comparison of reconstruction performance on real images is given in Fig.~\ref{compare_SCface}. We can see that the face priors-based methods reconstruct unsatisfactory results. The reason may be that estimating accurate priors from real-world LR face images is a difficult problem. Meanwhile, inaccurate prior information will bring misleading guidance to the reconstruction process. In comparison, benefit from the CNN-Transformer Cooperation mechanism, which is the prominent difference between CTCNet and other methods, our CTCNet can recover cleaner facial details and faithful facial structures. We also verify the superiority of our CTCNet over the performance of downstream tasks such as face matching. The high-definition frontal face images of the test candidates are selected as the source samples, while the corresponding LR face images captured by the surveillance camera are treated as the target samples. To make the experiments more convincing, we conducted 10 cases. In each case, we randomly select five pairs of candidate samples and calculate the average similarity. The quantitative results can be seen in Table~\ref{Average_Similarity}. We can see that our method can achieve higher similarity in each case, which further indicates that our CTCNet can also produce more faithful HR faces in real-world surveillance scenarios, making it highly practical and applicable.

\subsection{Model Complexity Analysis}
\label{sec45}

As can be seen from the previous results, our model achieves better performance than most of the competitive methods in terms of quantitative and qualitative comparisons. In addition, the model size and execution time are also important indicators to measure the efficiency of the model. In Fig.~\ref{complexity_CelebA}, we provide a comparison with other models between parameter quantity, model performance, and execution time. Obviously, our CTCNet achieves the best quantitative results under the premise of comparable execution time and parameters. As a whole, our CTCNet achieves a better balance between model size, model performance, and execution time.

\section{Conclusions}

In this work, we proposed a novel CNN-Transformer Cooperation Network (CTCNet) for face super-resolution. CTCNet uses the multi-scale connected encoder-decoder architecture as the backbone and exhibits extraordinary results. Specifically, we designed an efficient Local-Global Feature Cooperation Module (LGCM), which consists of a Facial Structure Attention Unit (FSAU) and a Transformer block, to focus on local facial details and global facial structures simultaneously. Meanwhile, to further improve the restoration results, we presented a Multi-scale Feature Fusion Unit (MFFU) to adaptively and elaborately fuse the features from different scales and depths. 
Extensive experiments on both simulated and real-world datasets have demonstrated the superiority of CTCNet over some competitive methods in terms of quantitative and qualitative comparisons. Furthermore, its reconstructed images show excellent results in downstream tasks such as face matching, which fully demonstrates its practicality and applicability.

\bibliographystyle{IEEEtran}
\bibliography{reference}

\begin{thebibliography}{10}
\providecommand{\url}[1]{#1}
\csname url@samestyle\endcsname
\providecommand{\newblock}{\relax}
\providecommand{\bibinfo}[2]{#2}
\providecommand{\BIBentrySTDinterwordspacing}{\spaceskip=0pt\relax}
\providecommand{\BIBentryALTinterwordstretchfactor}{4}
\providecommand{\BIBentryALTinterwordspacing}{\spaceskip=\fontdimen2\font plus
\BIBentryALTinterwordstretchfactor\fontdimen3\font minus
  \fontdimen4\font\relax}
\providecommand{\BIBforeignlanguage}[2]{{%
\expandafter\ifx\csname l@#1\endcsname\relax
\typeout{** WARNING: IEEEtran.bst: No hyphenation pattern has been}%
\typeout{** loaded for the language `#1'. Using the pattern for}%
\typeout{** the default language instead.}%
\else
\language=\csname l@#1\endcsname
\fi
#2}}
\providecommand{\BIBdecl}{\relax}
\BIBdecl

\bibitem{ma2020deep}
C.~Ma, Z.~Jiang, Y.~Rao, J.~Lu, and J.~Zhou, ``Deep face super-resolution with
  iterative collaboration between attentive recovery and landmark estimation,''
  in \emph{Proceedings of the IEEE Conference on Computer Vision and Pattern
  Recognition}, 2020, pp. 5569--5578.

\bibitem{hu2020face}
X.~Hu, W.~Ren, J.~LaMaster, X.~Cao, X.~Li, Z.~Li, B.~Menze, and W.~Liu, ``Face
  super-resolution guided by 3d facial priors,'' in \emph{Proceedings of the
  European Conference on Computer Vision}, 2020, pp. 763--780.

\bibitem{cai2019fcsr}
J.~Cai, H.~Han, S.~Shan, and X.~Chen, ``Fcsr-gan: Joint face completion and
  super-resolution via multi-task learning,'' \emph{IEEE Transactions on
  Biometrics, Behavior, and Identity Science}, vol.~2, no.~2, pp. 109--121,
  2019.

\bibitem{chen2018fsrnet}
Y.~Chen, Y.~Tai, X.~Liu, C.~Shen, and J.~Yang, ``Fsrnet: End-to-end learning
  face super-resolution with facial priors,'' in \emph{Proceedings of the IEEE
  Conference on Computer Vision and Pattern Recognition}, 2018, pp. 2492--2501.

\bibitem{xin2020facial}
J.~Xin, N.~Wang, X.~Jiang, J.~Li, X.~Gao, and Z.~Li, ``Facial attribute
  capsules for noise face super resolution,'' in \emph{Proceedings of the AAAI
  Conference on Artificial Intelligence}, 2020, pp. 12\,476--12\,483.

\bibitem{zhang2018super}
K.~Zhang, Z.~Zhang, C.-W. Cheng, W.~H. Hsu, Y.~Qiao, W.~Liu, and T.~Zhang,
  ``Super-identity convolutional neural network for face hallucination,'' in
  \emph{Proceedings of the European Conference on Computer Vision}, 2018, pp.
  183--198.

\bibitem{kim2019progressive}
D.~Kim, M.~Kim, G.~Kwon, and D.-S. Kim, ``Progressive face super-resolution via
  attention to facial landmark,'' \emph{arXiv preprint arXiv:1908.08239}, 2019.

\bibitem{chen2020learning}
C.~Chen, D.~Gong, H.~Wang, Z.~Li, and K.-Y.~K. Wong, ``Learning spatial
  attention for face super-resolution,'' \emph{IEEE Transactions on Image
  Processing}, vol.~30, pp. 1219--1231, 2021.

\bibitem{yang2021gan}
T.~Yang, P.~Ren, X.~Xie, and L.~Zhang, ``Gan prior embedded network for blind
  face restoration in the wild,'' in \emph{Proceedings of the IEEE Conference
  on Computer Vision and Pattern Recognition}, 2021, pp. 672--681.

\bibitem{wang2022restoreformer}
Z.~Wang, J.~Zhang, R.~Chen, W.~Wang, and P.~Luo, ``Restoreformer: High-quality
  blind face restoration from undegraded key-value pairs,'' in
  \emph{Proceedings of the IEEE Conference on Computer Vision and Pattern
  Recognition}, 2022, pp. 17\,512--17\,521.

\bibitem{chen2020robust}
L.~Chen, J.~Pan, J.~Jiang, J.~Zhang, and Y.~Wu, ``Robust face super-resolution
  via position relation model based on global face context,'' \emph{IEEE
  Transactions on Image Processing}, vol.~29, pp. 9002--9016, 2020.

\bibitem{gao2021constructing}
G.~Gao, Y.~Yu, J.~Xie, J.~Yang, M.~Yang, and J.~Zhang, ``Constructing
  multilayer locality-constrained matrix regression framework for noise robust
  face super-resolution,'' \emph{Pattern Recognition}, vol. 110, p. 107539,
  2021.

\bibitem{zhang2022pro}
Y.~Zhang, X.~Yu, X.~Lu, and P.~Liu, ``Pro-uigan: Progressive face hallucination
  from occluded thumbnails,'' \emph{IEEE Transactions on Image Processing},
  vol.~31, pp. 3236--3250, 2022.

\bibitem{wang2020deep}
Z.~Wang, J.~Chen, and S.~C. Hoi, ``Deep learning for image super-resolution: A
  survey,'' \emph{IEEE Transactions on Pattern Analysis and Machine
  Intelligence}, vol.~43, no.~10, pp. 3365--3387, 2020.

\bibitem{li2021beginner}
J.~Li, Z.~Pei, and T.~Zeng, ``From beginner to master: A survey for deep
  learning-based single-image super-resolution,'' \emph{arXiv preprint
  arXiv:2109.14335}, 2021.

\bibitem{gao2022lightweight}
G.~Gao, Z.~Wang, J.~Li, W.~Li, Y.~Yu, and T.~Zeng, ``Lightweight bimodal
  network for single-image super-resolution via symmetric cnn and recursive
  transformer,'' in \emph{Proceedings of the International Joint Conference on
  Artificial Intelligence}, 2022, pp. 661--669.

\bibitem{MSRN}
J.~Li, F.~Fang, K.~Mei, and G.~Zhang, ``Multi-scale residual network for image
  super-resolution,'' in \emph{Proceedings of the European Conference on
  Computer Vision (ECCV)}, 2018, pp. 517--532.

\bibitem{DRN}
Y.~Guo, J.~Chen, J.~Wang, Q.~Chen, J.~Cao, Z.~Deng, Y.~Xu, and M.~Tan,
  ``Closed-loop matters: Dual regression networks for single image
  super-resolution,'' in \emph{Proceedings of the IEEE Conference on Computer
  Vision and Pattern Recognition (CVPR)}, 2020, pp. 5407--5416.

\bibitem{GLADSR}
X.~Zhang, P.~Gao, S.~Liu, K.~Zhao, G.~Li, L.~Yin, and C.~W. Chen, ``Accurate
  and efficient image super-resolution via global-local adjusting dense
  network,'' \emph{IEEE Transactions on multimedia}, vol.~23, pp. 1924--1937,
  2021.

\bibitem{gao2022feature}
G.~Gao, W.~Li, J.~Li, F.~Wu, H.~Lu, and Y.~Yu, ``Feature distillation
  interaction weighting network for lightweight image super-resolution,'' in
  \emph{Proceedings of the AAAI Conference on Artificial Intelligence},
  vol.~36, no.~1, 2022, pp. 661--669.

\bibitem{lu2021face}
T.~Lu, Y.~Wang, Y.~Zhang, Y.~Wang, L.~Wei, Z.~Wang, and J.~Jiang, ``Face
  hallucination via split-attention in split-attention network,'' in
  \emph{Proceedings of the 29th ACM International Conference on Multimedia},
  2021, pp. 5501--5509.

\bibitem{li2020learning}
M.~Li, Z.~Zhang, J.~Yu, and C.~W. Chen, ``Learning face image super-resolution
  through facial semantic attribute transformation and self-attentive structure
  enhancement,'' \emph{IEEE Transactions on Multimedia}, vol.~23, pp. 468--483,
  2021.

\bibitem{hu2018squeeze}
J.~Hu, L.~Shen, and G.~Sun, ``Squeeze-and-excitation networks,'' in
  \emph{Proceedings of the IEEE Conference on Computer Vision and Pattern
  Recognition}, 2018, pp. 7132--7141.

\bibitem{wang2020eca}
Q.~Wang, B.~Wu, P.~Zhu, P.~Li, W.~Zuo, and Q.~Hu, ``Eca-net: Efficient channel
  attention for deep convolutional neural networks,'' in \emph{Proceedings of
  the IEEE Conference on Computer Vision and Pattern Recognition}, 2020, pp.
  11\,534--11\,542.

\bibitem{zhang2018image}
Y.~Zhang, K.~Li, K.~Li, L.~Wang, B.~Zhong, and Y.~Fu, ``Image super-resolution
  using very deep residual channel attention networks,'' in \emph{Proceedings
  of the European Conference on Computer Vision}, 2018, pp. 286--301.

\bibitem{dai2019second}
T.~Dai, J.~Cai, Y.~Zhang, S.-T. Xia, and L.~Zhang, ``Second-order attention
  network for single image super-resolution,'' in \emph{Proceedings of the IEEE
  Conference on Computer Vision and Pattern Recognition}, 2019, pp.
  11\,065--11\,074.

\bibitem{niu2020single}
B.~Niu, W.~Wen, W.~Ren, X.~Zhang, L.~Yang, S.~Wang, K.~Zhang, X.~Cao, and
  H.~Shen, ``Single image super-resolution via a holistic attention network,''
  in \emph{Proceedings of the European Conference on Computer Vision}, 2020,
  pp. 191--207.

\bibitem{vaswani2017attention}
A.~Vaswani, N.~Shazeer, N.~Parmar, J.~Uszkoreit, L.~Jones, A.~N. Gomez,
  {\L}.~Kaiser, and I.~Polosukhin, ``Attention is all you need,'' in
  \emph{Proceedings of the Advances in Neural Information Processing Systems},
  2017, pp. 5998--6008.

\bibitem{devlin2018bert}
J.~Devlin, M.-W. Chang, K.~Lee, and K.~Toutanova, ``Bert: Pre-training of deep
  bidirectional transformers for language understanding,'' \emph{arXiv preprint
  arXiv:1810.04805}, 2018.

\bibitem{dosovitskiy2020image}
A.~Dosovitskiy, L.~Beyer, A.~Kolesnikov, D.~Weissenborn, X.~Zhai,
  T.~Unterthiner, M.~Dehghani, M.~Minderer, G.~Heigold, S.~Gelly \emph{et~al.},
  ``An image is worth 16x16 words: Transformers for image recognition at
  scale,'' \emph{arXiv preprint arXiv:2010.11929}, 2020.

\bibitem{touvron2021training}
H.~Touvron, M.~Cord, M.~Douze, F.~Massa, A.~Sablayrolles, and H.~J{\'e}gou,
  ``Training data-efficient image transformers \& distillation through
  attention,'' in \emph{Proceedings of the International Conference on Machine
  Learning}, 2021, pp. 10\,347--10\,357.

\bibitem{carion2020end}
N.~Carion, F.~Massa, G.~Synnaeve, N.~Usunier, A.~Kirillov, and S.~Zagoruyko,
  ``End-to-end object detection with transformers,'' in \emph{Proceedings of
  the European Conference on Computer Vision}, 2020, pp. 213--229.

\bibitem{zhu2020deformable}
X.~Zhu, W.~Su, L.~Lu, B.~Li, X.~Wang, and J.~Dai, ``Deformable detr: Deformable
  transformers for end-to-end object detection,'' \emph{arXiv preprint
  arXiv:2010.04159}, 2020.

\bibitem{liang2021swinir}
J.~Liang, J.~Cao, G.~Sun, K.~Zhang, L.~Van~Gool, and R.~Timofte, ``Swinir:
  Image restoration using swin transformer,'' in \emph{Proceedings of the
  IEEE/CVF International Conference on Computer Vision}, 2021, pp. 1833--1844.

\bibitem{lu2021efficient}
Z.~Lu, H.~Liu, J.~Li, and L.~Zhang, ``Efficient transformer for single image
  super-resolution,'' \emph{arXiv preprint arXiv:2108.11084}, 2021.

\bibitem{esser2021taming}
P.~Esser, R.~Rombach, and B.~Ommer, ``Taming transformers for high-resolution
  image synthesis,'' in \emph{Proceedings of the IEEE/CVF Conference on
  Computer Vision and Pattern Recognition}, 2021, pp. 12\,873--12\,883.

\bibitem{wang2021uformer}
Z.~Wang, X.~Cun, J.~Bao, and J.~Liu, ``Uformer: A general u-shaped transformer
  for image restoration,'' in \emph{Proceedings of the IEEE/CVF Conference on
  Computer Vision and Pattern Recognition (CVPR)}, 2022, pp. 17\,683--17\,693.

\bibitem{zamir2021restormer}
S.~W. Zamir, A.~Arora, S.~Khan, M.~Hayat, F.~S. Khan, and M.-H. Yang,
  ``Restormer: Efficient transformer for high-resolution image restoration,''
  in \emph{Proceedings of the IEEE/CVF Conference on Computer Vision and
  Pattern Recognition (CVPR)}, 2022, pp. 5728--5739.

\bibitem{chen2017adversarial}
Y.~Chen, C.~Shen, X.-S. Wei, L.~Liu, and J.~Yang, ``Adversarial posenet: A
  structure-aware convolutional network for human pose estimation,'' in
  \emph{Proceedings of the IEEE International Conference on Computer Vision},
  2017, pp. 1212--1221.

\bibitem{newell2016stacked}
A.~Newell, K.~Yang, and J.~Deng, ``Stacked hourglass networks for human pose
  estimation,'' in \emph{Proceedings of the European Conference on Computer
  Vision}, 2016, pp. 483--499.

\bibitem{he2016deep}
K.~He, X.~Zhang, S.~Ren, and J.~Sun, ``Deep residual learning for image
  recognition,'' in \emph{Proceedings of the IEEE Conference on Computer Vision
  and Pattern Recognition}, 2016, pp. 770--778.

\bibitem{Dan2016gauss}
D.~Hendrycks and K.~Gimpel, ``Gaussian error linear units (gelus),''
  \emph{arXiv preprint arXiv:1606.08415}, 2016.

\bibitem{ledig2017photo}
C.~Ledig, L.~Theis, F.~Husz{\'a}r, J.~Caballero, A.~Cunningham, A.~Acosta,
  A.~Aitken, A.~Tejani, J.~Totz, Z.~Wang \emph{et~al.}, ``Photo-realistic
  single image super-resolution using a generative adversarial network,'' in
  \emph{Proceedings of the IEEE Conference on Computer Vision and Pattern
  Recognition}, 2017, pp. 4681--4690.

\bibitem{wang2018esrgan}
X.~Wang, K.~Yu, S.~Wu, J.~Gu, Y.~Liu, C.~Dong, Y.~Qiao, and C.~Change~Loy,
  ``Esrgan: Enhanced super-resolution generative adversarial networks,'' in
  \emph{Proceedings of the European Conference on Computer Vision (ECCV)
  workshops}, 2018, pp. 1--16.

\bibitem{isola2017image}
P.~Isola, J.-Y. Zhu, T.~Zhou, and A.~A. Efros, ``Image-to-image translation
  with conditional adversarial networks,'' in \emph{Proceedings of the IEEE
  Conference on Computer Vision and Pattern Recognition}, 2017, pp. 1125--1134.

\bibitem{simonyan2014very}
K.~Simonyan and A.~Zisserman, ``Very deep convolutional networks for
  large-scale image recognition,'' \emph{arXiv preprint arXiv:1409.1556}, 2014.

\bibitem{liu2015deep}
Z.~Liu, P.~Luo, X.~Wang, and X.~Tang, ``Deep learning face attributes in the
  wild,'' in \emph{Proceedings of the IEEE International Conference on Computer
  Vision}, 2015, pp. 3730--3738.

\bibitem{le2012interactive}
V.~Le, J.~Brandt, Z.~Lin, L.~Bourdev, and T.~S. Huang, ``Interactive facial
  feature localization,'' in \emph{Proceedings of the European Conference on
  Computer Vision}, 2012, pp. 679--692.

\bibitem{grgic2011scface}
M.~Grgic, K.~Delac, and S.~Grgic, ``Scface--surveillance cameras face
  database,'' \emph{Multimedia Tools and Applications}, vol.~51, no.~3, pp.
  863--879, 2011.

\bibitem{wang2004image}
Z.~Wang, A.~C. Bovik, H.~R. Sheikh, and E.~P. Simoncelli, ``Image quality
  assessment: from error visibility to structural similarity,'' \emph{IEEE
  Transactions on Image Processing}, vol.~13, no.~4, pp. 600--612, 2004.

\bibitem{zhang2018unreasonable}
R.~Zhang, P.~Isola, A.~A. Efros, E.~Shechtman, and O.~Wang, ``The unreasonable
  effectiveness of deep features as a perceptual metric,'' in \emph{Proceedings
  of the IEEE Conference on Computer Vision and Pattern Recognition}, 2018, pp.
  586--595.

\bibitem{sheikh2006image}
H.~R. Sheikh and A.~C. Bovik, ``Image information and visual quality,''
  \emph{IEEE Transactions on Image Processing}, vol.~15, no.~2, pp. 430--444,
  2006.

\bibitem{obukhov2020quality}
A.~Obukhov and M.~Krasnyanskiy, ``Quality assessment method for gan based on
  modified metrics inception score and fr{\'e}chet inception distance,'' in
  \emph{Proceedings of the Computational Methods in Systems and Software},
  2020, pp. 102--114.

\end{thebibliography}

%

\begin{IEEEbiography}[{\includegraphics[width=1in,height=1.25in,clip,keepaspectratio]{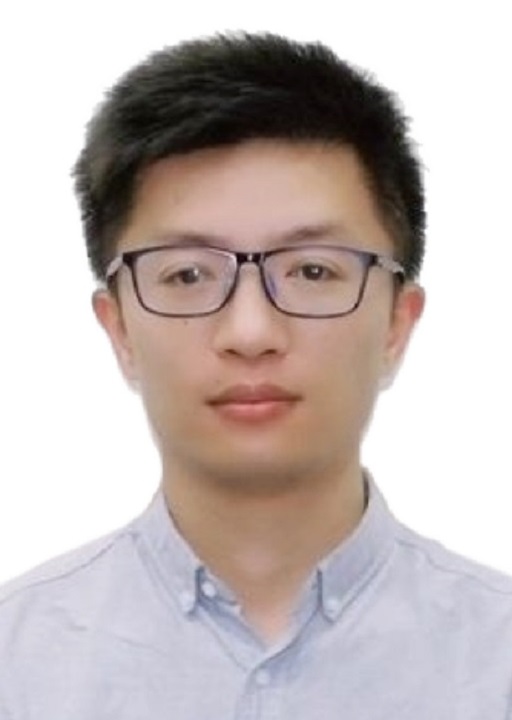}}]{Guangwei Gao}
(Senior Member, IEEE) received the Ph.D. degree in pattern recognition and intelligence systems from the Nanjing University of Science and Technology, Nanjing, in 2014. He was a Visiting Student of the Department of Computing, The Hong Kong Polytechnic University, in 2011 and 2013, respectively. He was also a Project Researcher with the National Institute of Informatics, Japan, in 2019. He is currently an Associate Professor at Nanjing University of Posts and Telecommunications. His research interests include pattern recognition and computer vision. He has published more than 60 scientific papers in IEEE TIP/TCSVT/TITS/TMM/TIFS, ACM TOIT/TOMM, PR, AAAI, IJCAI, etc. Personal website: \textit{https://guangweigao.github.io}.
\end{IEEEbiography}

\begin{IEEEbiography}[{\includegraphics[width=1in,height=1.25in,clip,keepaspectratio]{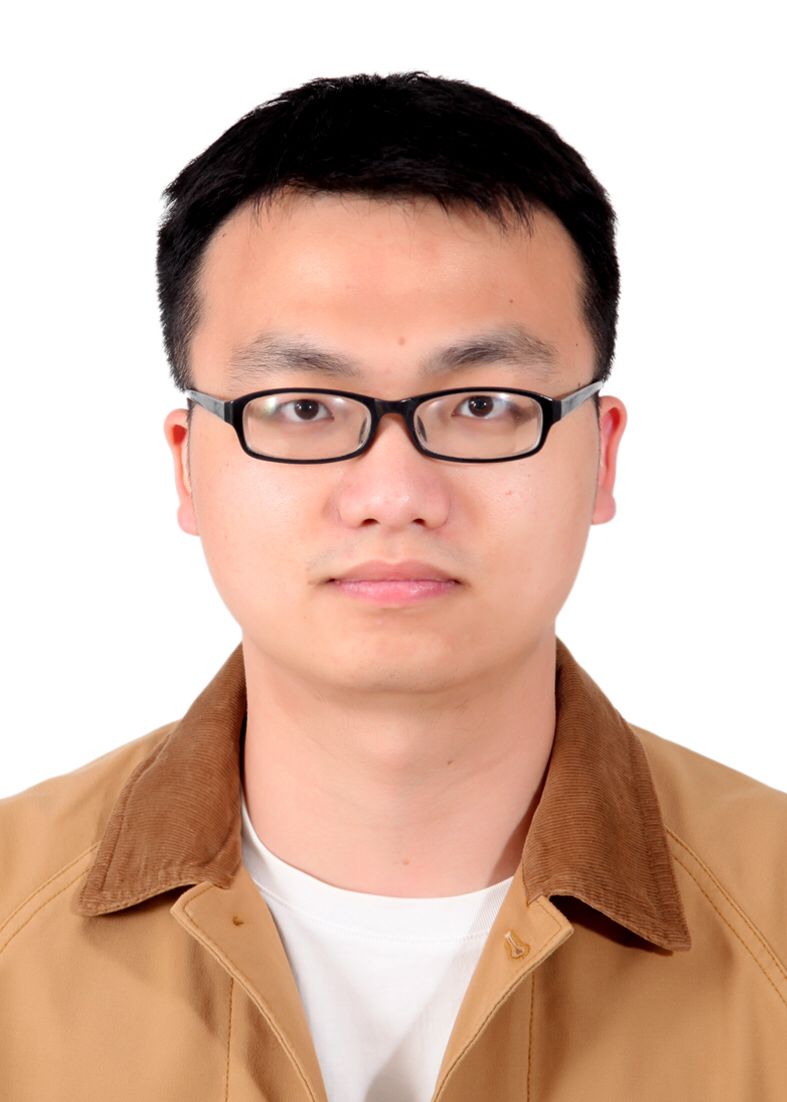}}]{Zixiang Xu}
 received the B.S. degree in Communication Engineering from Tongda College of Nanjing University of Posts and Telecommunications, Jiangsu, China, in 2020. He is currently pursuing the M.S. degree with the College of Automation \& College of Artificial Intelligence, Nanjing University of Posts and Telecommunications. His research interests include image super-resolution.
\end{IEEEbiography}

\begin{IEEEbiography}[{\includegraphics[width=1in,height=1.25in,clip,keepaspectratio]{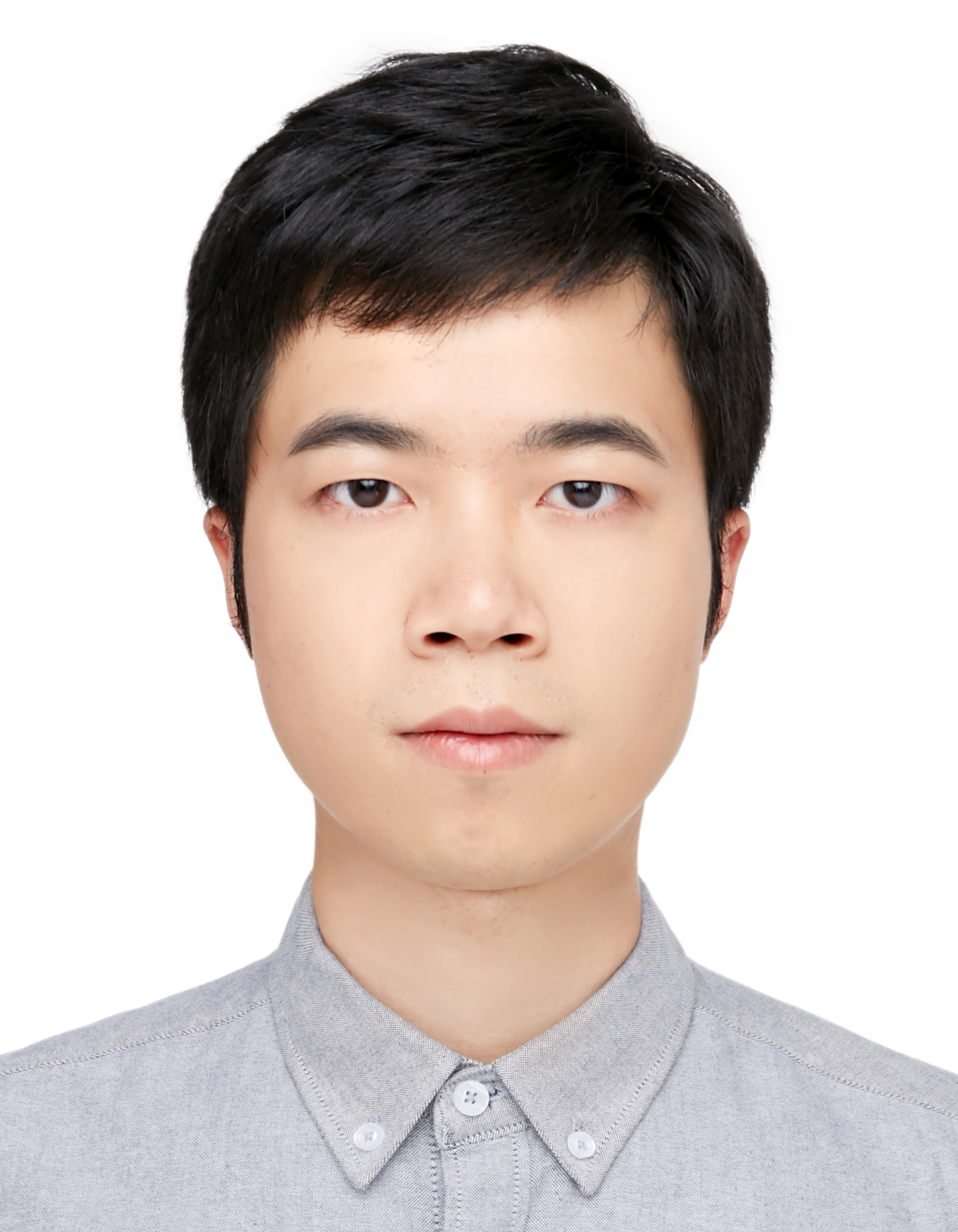}}]{Juncheng Li}
received the Ph.D. degree in Computer Science and Technology from East China Normal University, in 2021, and was a Postdoctoral Fellow at the Center for Mathematical Artificial Intelligence (CMAI), The Chinese University of Hong Kong. He is currently an Assistant Professor at the School of Communication \& Information Engineering, Shanghai University. His main research interests include image restoration, computer vision, and medical image processing. He has published more than 27 scientific papers in IEEE TIP, IEEE TNNLS, IEEE TMM, ICCV, ECCV, AAAI, and IJCAI.
\end{IEEEbiography}

\begin{IEEEbiography}[{\includegraphics[width=1in,height=1.25in,clip,keepaspectratio]{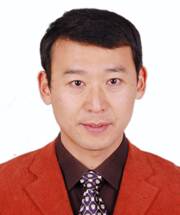}}]{Jian Yang}
(Member, IEEE) received the Ph.D. degree from the Nanjing University of Science and Technology (NJUST), Nanjing, China. From 2006 to 2007, he was a postdoctoral fellow with the Department of Computer Science, New Jersey Institute of Technology. He is currently a professor with the School of Computer Science and Technology, NJUST. He is the author of more than 400 scientific papers in pattern recognition and computer vision. His papers have been cited more than 28000 times in the Scholar Google. His research interests include pattern recognition, computer vision, and machine learning. He is/was an associate editor for Pattern Recognition and IEEE Transactions Neural Networks and Learning Systems. He is a fellow of IAPR.
\end{IEEEbiography}

\begin{IEEEbiography}[{\includegraphics[width=1in,height=1.25in,clip,keepaspectratio]{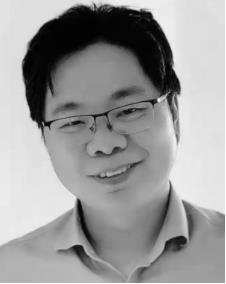}}]{Tieyong Zeng}
received the B.S. degree from Peking University, Beijing, China, the M.S. degree
from Ecole Polytechnique, Palaiseau, France, and the Ph.D. degree from the University of Paris XIII, Paris, France, in 2000, 2004, and 2007, respectively. He is currently a professor with the Department of Mathematics, The Chinese University of Hong Kong (CUHK). Together with colleagues, he has founded the Center for Mathematical Artificial Intelligence (CMAI) since 2020 and served as the director of CMAI. His research interests include image processing, optimization, artificial intelligence, scientific computing, computer vision, machine learning, and inverse problems. He has published around 100 papers in prestigious journals such as SIAM Journal on Imaging Sciences, SIAM Journal on Scientific Computing, Journal of Scientific Computing, IEEE Transactions on Pattern Analysis and Machine Intelligence (TPAMI), International Journal of Computer Vision (IJCV), IEEE Transactions on Neural Networks and Learning Systems (TNNLS), IEEE Transactions on Image Processing (TIP), IEEE Medical Imaging (TMI), and Pattern Recognition.
\end{IEEEbiography}

\begin{IEEEbiography}[{\includegraphics[width=1in,height=1.25in,clip,keepaspectratio]{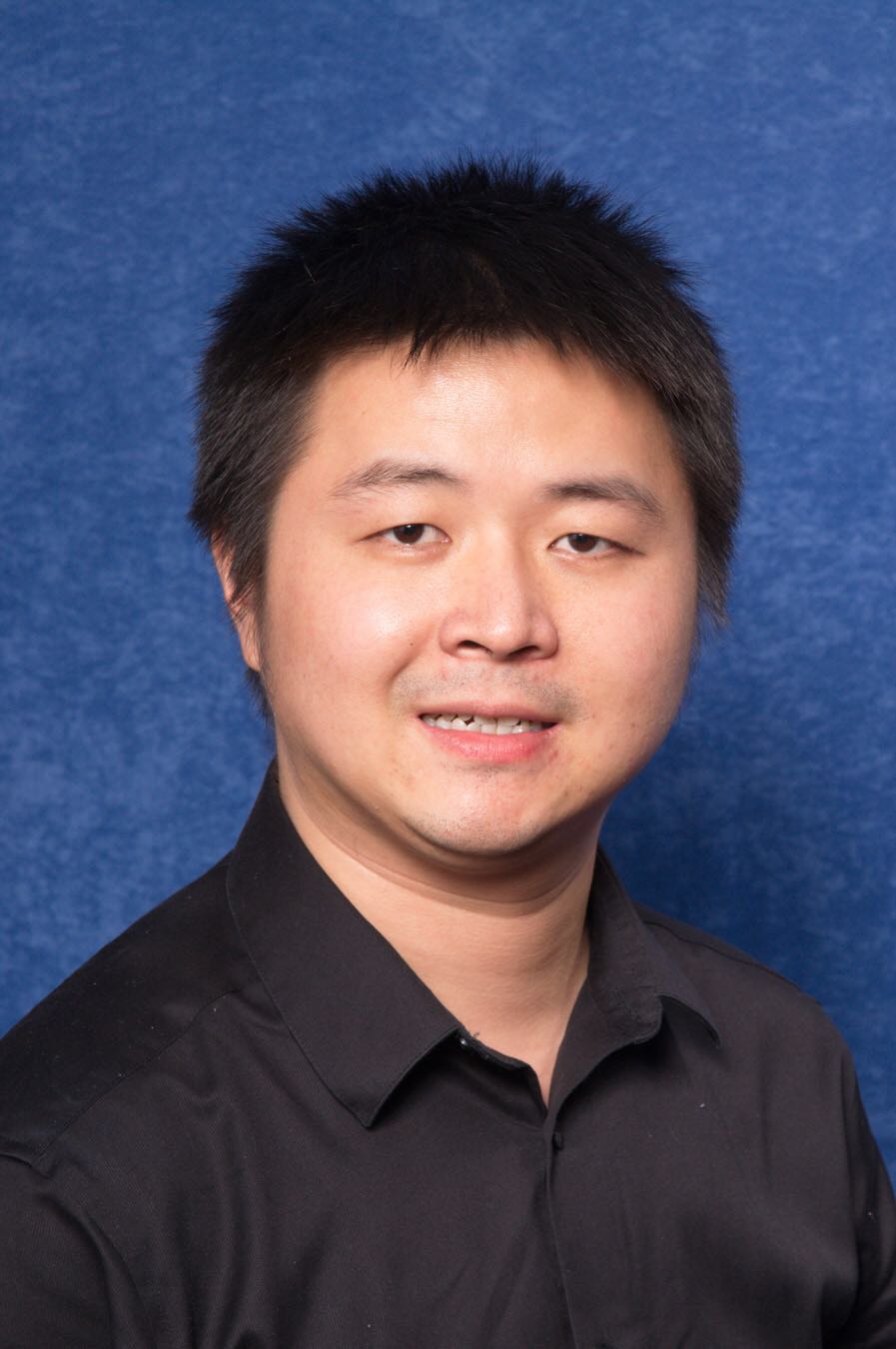}}]{Guo-Jun Qi}
(Fellow, IEEE) is the chief scientist and the director of the Seattle Research Center in the OPPO Research USA since 2021. Before that, he was the chief scientist who led and oversaw an international R\&D team in the domain of multiple intelligent cloud services, including smart cities, visual computing services, medical intelligent services, and connected vehicle services at Futurewei since 2018. He was a faculty member in the Department of Computer Science and the director of MAchine Perception and LEarning (MAPLE) Lab at the University of Central Florida, Orlando, Florida since 2014. Prior to that, he was also a research staff member at IBM T.J. Watson Research Center, Yorktown Heights, New York.
\end{IEEEbiography}





\end{document}